\documentclass{robocoach_arxiv}

\usepackage[utf8]{inputenc}
\usepackage[T1]{fontenc}
\usepackage{url}
\usepackage{booktabs}
\usepackage{amsmath,amssymb,mathtools}
\usepackage{graphicx}
\usepackage{xcolor}
\usepackage{multirow}
\usepackage{enumitem}
\usepackage{algorithm}
\usepackage{algorithmic}
\usepackage{placeins}
\usepackage{flafter}
\usepackage{wrapfig}

\newcommand{\PaperTitlePlain}{%
  RoboCoach: World Models as Active Coaches for Compositional Robot Skills%
}

\newcommand{\methodname}{\textsc{RoboCoach}}
\newcommand{\modelname}{\textsc{CoachWorld}}
\newcommand{\wm}{\mathcal{W}}
\newcommand{\pol}{\pi}
\newcommand{\backbone}{\pi_{\mathrm{base}}}
\newcommand{\router}{\rho}

\newcommand{\experts}{\Pi}

\newcommand{\PaperTeaserCaption}{%
\textbf{Overview of \methodname{}.}
The Route--Imagine--Diagnose--Improve (RIDI) loop routes the next unfinished subtask to a reusable expert, rolls it forward in \modelname{}, diagnoses the first unresolved subtask, and converts recurring imagined failures into targeted demonstrations and expert updates.
Panels (A)--(D) correspond to Route (Sec.~\ref{sec:routing}), Imagine (Sec.~\ref{sec:coachworld}), Diagnose (Sec.~\ref{sec:diagnosis}), and Improve (Sec.~\ref{sec:allocation}).
}

\newcommand{\acquisitioncell}[1]{%
  \parbox[t]{\linewidth}{\raggedright #1\strut}%
}

\title{RoboCoach: World Models as Active Coaches\\for Compositional Robot Skills}
\runningtitle{RoboCoach}
\titlelogos{%
  \includegraphics[height=12mm]{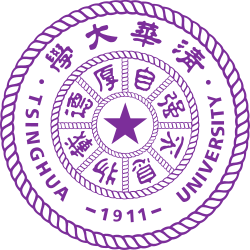}%
  \hspace{4mm}%
  \includegraphics[height=12mm]{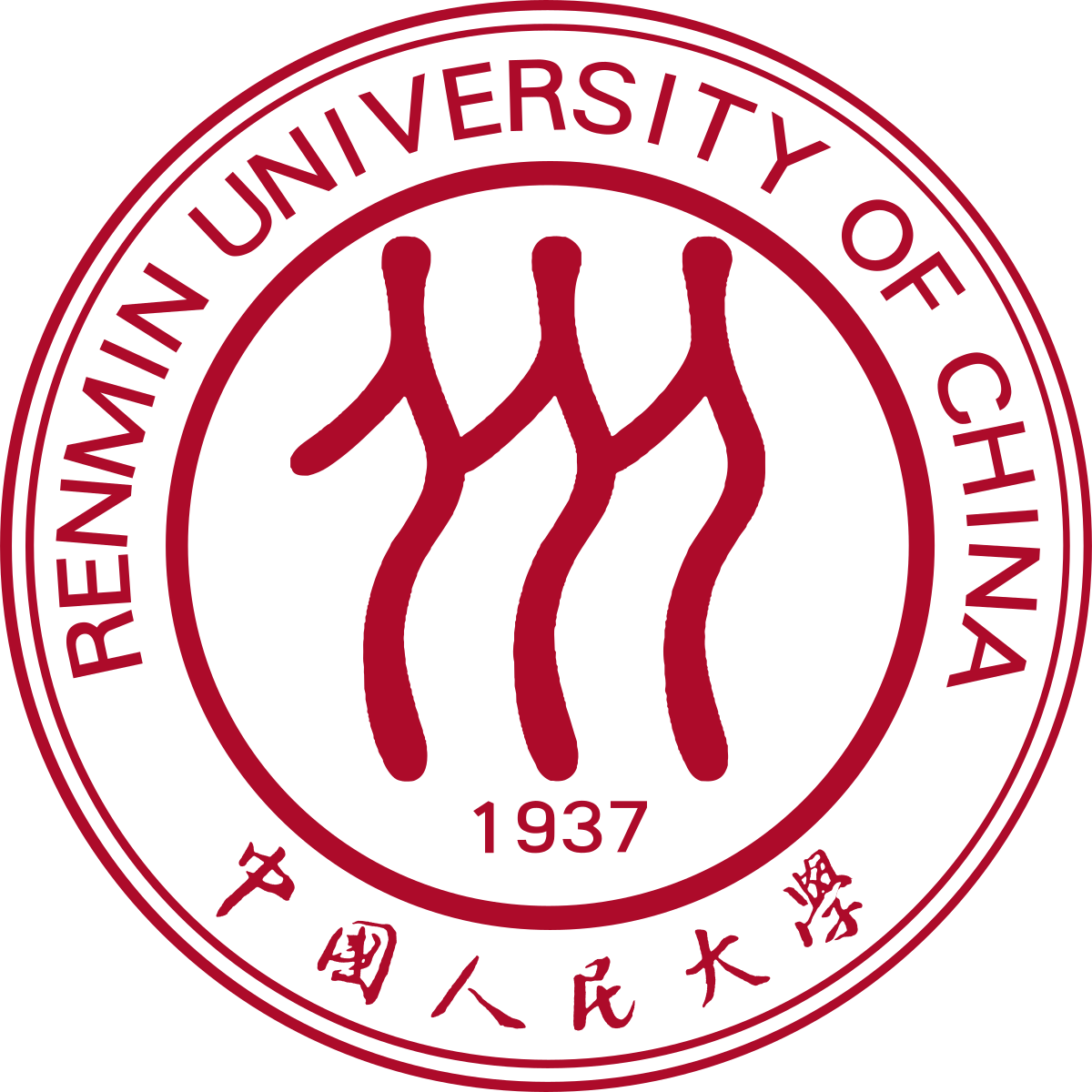}%
  \hfill
  \includegraphics[height=12mm]{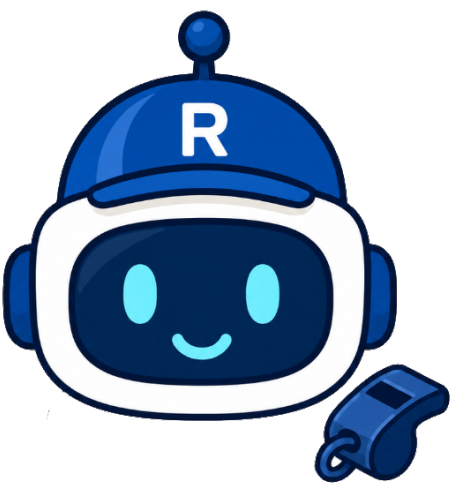}%
}
\titlefigure{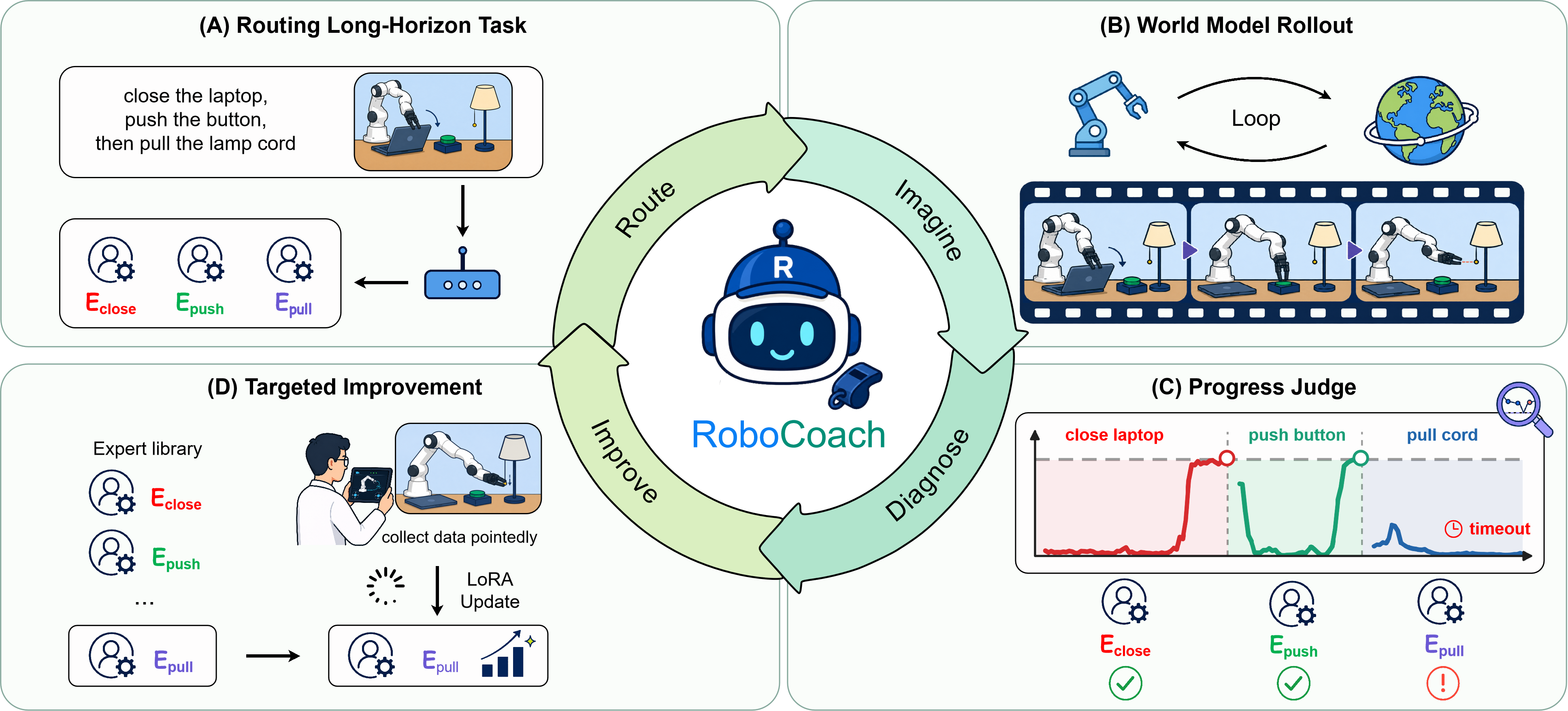}
\titlefigurecaption{\PaperTeaserCaption}
\titlefigurelabel{fig:teaser}
\projectlinks{\faGlobe\hspace{0.5em}Project page: {\urlstyle{tt}\url{https://robocoach-ai.github.io/}}}
\hypersetup{pdftitle={\PaperTitlePlain},
  pdfauthor={Jiajun Liu, Yifan Chen, Yichao Liu, Jiayi Zhang, Ruoqu Chen, Shaoxuan Xie, Guocai Yao, Mengdi Xu, Sen Cui, Changshui Zhang}}

\author[1,2,5,*]{Jiajun Liu}
\author[2,*]{Yifan Chen}
\author[2,*]{Yichao Liu}
\author[3]{Jiayi Zhang}
\author[2,5]{Ruoqu Chen}
\authorrow
\author[4]{Shaoxuan Xie}
\author[4]{Guocai Yao}
\author[2,5,\dagger]{Mengdi Xu}
\author[2,4,\dagger, \ddagger]{Sen Cui}
\author[2,\dagger]{Changshui Zhang}

\affiliation[1]{Renmin University of China}
\affiliation[2]{Tsinghua University}
\affiliation[3]{University of Nottingham}
\affiliationrow
\affiliation[4]{Beijing Academy of Artificial Intelligence}
\affiliation[5]{Shanghai Qizhi Institute}

\contribution[*]{Equal contributions}
\contribution[\dagger]{Corresponding authors}
\contribution[\ddagger]{Project leader}

\abstract{Long-horizon robot manipulation reuses skills across many task compositions, but improving these compositions with additional end-to-end demonstrations is costly. A practical self-improving system must decide both \textit{what to teach next} and \textit{where to apply that supervision}. We present \methodname{}, a world-model-guided coaching framework that uses
imagined failures to guide demonstration requests and expert updates. Its Route--Imagine--Diagnose--Improve (RIDI) loop executes reusable skill experts inside \modelname{}, our shared action-conditioned world model, and uses a progress
judge to record the first subtask that fails to complete. Aggregated records select
which subtask demonstrations to acquire and which expert adapters to update.
Across two simulation suites and two real-robot platforms, imagined and
deployed success correlate over 22 task--policy pairs ($\rho=0.840$). Controlled comparisons show that our coaching method outperforms matched baselines under matched data budgets and update schedules.
With only 150 additional subtask demonstrations, success rises from $13.3\%$ to $75.0\%$ on Franka and from $40.0\%$ to $83.8\%$ on AgileX. The coached experts also transfer to four held-out compositions, achieving an average success of $35.0\%$, compared with $0\%$ for a shared-policy baseline updated with uniformly acquired demonstrations. Together, these results show that world models can serve as active coaches, turning imagined failures into targeted supervision for modular policy improvement.
}
\correspondence{\email{jedward.jiajun@gmail.com},
  {\{\href{mailto:xumd@tsinghua.edu.cn}{xumd},
  \href{mailto:cuis@tsinghua.edu.cn}{cuis},
  \href{mailto:zcs@tsinghua.edu.cn}{zcs}\}@tsinghua.edu.cn}}

\begin{document}
\maketitle

\clearpage
{\setcounter{tocdepth}{2}\small\tableofcontents}
\clearpage

% Main text of the RoboCoach preprint.
\section{Introduction}

An embodied self-improving system must decide not only how to act, but also what experience to acquire next. This decision is especially consequential in long-horizon manipulation, where tasks such as preparing tea or setting a table require a sequence of reusable skills. Each skill changes the state in which the next is executed, so a local execution error can propagate and cause the entire task to fail. Additional end-to-end demonstrations can improve the policy, but covering the long tail of all possible skill compositions is expensive and difficult to scale. The burden is particularly heavy on physical robots, where every trajectory consumes data-collection effort, hardware time, environment resets, and safety oversight. These constraints raise a concrete policy-improvement question: \textit{which skill should the next demonstrations target, and which policy component should receive the update, given that each additional trajectory is costly?} %

Recent advances in robot learning provide important capabilities for addressing these questions.
Generalist vision--language--action models (VLAs) and hierarchical systems support increasingly capable execution across tasks~\citep{kim24openvla,intelligence2025pi_,shi2025hi}, while active and corrective imitation learning use interaction feedback and policy failures to select additional supervision~\citep{hgdagger,mu2024adademo}.
Action-conditioned world models support policy evaluation and improvement through imagined interaction, allowing robots to examine action consequences before physical execution~\citep{guo2025ctrl,li2025worldevalworldmodelrealworld,yang2026rise,liu2026worldvlaloop}.
Meanwhile, modular policies provide task- or skill-specific components that can be adapted independently~\citep{luo2026coral,zhang2026atomicvlaunlockingpotentialatomic}.
The central challenge is to connect these capabilities into a learning process: predicted execution must yield a concrete request for real supervision, and that supervision must reach the policy component associated with the identified learning need.
For long-horizon tasks, this connection should operate at the level of reusable skills, so that additional demonstrations improve skills that transfer across tasks.

Our approach separates \emph{shared prediction} from \emph{selective adaptation}.
An action-conditioned world model learns regularities of motion, contact, and object-state change from diverse robot interactions, providing a shared predictive model across tasks and embodiments.
Execution, however, depends on task stage, morphology, and context, and is better
served by modular experts that specialize over time.
We therefore build execution around reusable skill experts on a shared VLA backbone,
so that each expert is an explicit target for adaptation.
Each expert specializes in an atomic skill, such as picking, placing, or pressing,
and can be reused across objects and task sequences.
A \emph{subtask--expert pair} thus links the behavior needing teaching to the reusable
component that learns from it.
This makes long-horizon compositionality useful for both execution and improvement: shared
prediction decides what to teach next, while selective adaptation updates the
corresponding reusable skill.
Together, they suggest a broader role for world models: they should not merely
simulate more experience, but should help decide where scarce real experience is
most valuable.

We instantiate this idea in \methodname{}, a world-model-guided active coaching framework organized as a Route--Imagine--Diagnose--Improve (RIDI) loop (Fig.~\ref{fig:teaser}).
The router selects the next unfinished subtask and dispatches its corresponding skill expert.
The expert then interacts with \modelname{}, our shared action-conditioned world model, in closed loop.
A progress judge reads the imagined observations to determine whether the active subtask has been completed and whether execution should advance to the next expert.
When a subtask remains incomplete at its time limit, the system records the active subtask--expert pair.
Across imagined trials, these records prioritize requests for new demonstrations.
The acquired demonstrations update only the corresponding expert LoRA adapters~\citep{hu2021lora}. Updated experts then return to the route, closing the RIDI loop.

We evaluate \methodname{} on LIBERO and RoboTwin, and on Franka and AgileX robots.
Our evaluation examines the evidence underlying coaching through video-prediction fidelity, agreement with deployed policy outcomes, and progress-judge reliability.
Across 22 task--policy pairs, imagined and deployed success rates achieve a Spearman correlation of $\rho=0.840$.
Controlled comparisons demonstrate the value of coupling acquisition with adaptation: applying the same targeted demonstrations to corresponding skill experts rather than a shared global adapter improves final complete-task success by $3.4$ and $13.2$ pp on LIBERO and RoboTwin, respectively.
On real robots, $150$ additional subtask demonstrations per platform raise success from $13.3\%$ to $75.0\%$ on Franka and from $40.0\%$ to $83.8\%$ on AgileX, whereas uniform acquisition with a shared adapter reaches $30.0\%$ and $47.5\%$ under the same budgets.
The coached experts achieve $35.0\%$ average success on four held-out compositions spanning longer continuations, cross-task combinations, and reordered skills, compared with $0\%$ for the shared-policy baseline.

We summarize our key contributions as follows:
\textbf{First,} we study embodied self-improvement through coupled decisions about \textit{which demonstrations to acquire} and \textit{which policy components to update}. We instantiate these decisions with subtask--expert pairs, making reusable skills explicit targets for both data acquisition and adaptation.
\textbf{Second,} we introduce \methodname{}, whose RIDI loop connects \modelname{}, progress-based diagnosis, and selective expert adaptation to turn imagined failures into targeted supervision and iterative policy improvement.
\textbf{Third,} our experiments demonstrate substantial gains with limited additional supervision, reveal the benefit of directing identical acquired demonstrations to corresponding experts, and show that coached skills can be recombined beyond the task sequences used for improvement.

\section{Related Work}

\paragraph{Generalist VLAs for long-horizon manipulation.}
Generalist VLAs learn transferable visuomotor priors from large robot datasets and vision--language pretraining~\citep{rt2,kim24openvla,octo_2023,black2024pi_0}.
RL post-training further improves execution robustness through interaction feedback~\citep{li2025simplevla,lu2025vla,xiao2025worldenvleveragingworldmodel}.
\methodname{} uses reusable skill experts as explicit targets for additional supervision, linking the subtask to be demonstrated with the policy component that receives its update.

\paragraph{World models for embodied AI.}
Action-conditioned world models simulate future observations for imagined rollout and policy evaluation~\citep{hafner2025dreamerv3,li2025worldevalworldmodelrealworld,quevedo2025worldgymworldmodelenvironment,li2026dworldeval,tseng2026sc3}.
They also support policy optimization through imagined interaction~\citep{zhou2024robodreamer,WMPO2025,jiang2026wovr,yang2026rise} and generate additional supervision for robot policies~\citep{liu2026worldvlaloop,gigaai2025gigaworld0,nvidia2025worldsimulationvideofoundation}.
Recent models improve action controllability by projecting end-effector trajectories, kinematic structures, or contact fields into the camera view~\citep{jiang2025enerverseac,oscar2026,eawm2026,huang2026paiworld3dconsistentworldfoundation,xiang2026geometry,huang2026a2world}.
By exploiting this predictive capability, we developed world model's additional function: localize the bottleneck skill expert and allocate demonstrations~\citep{li2025comprehensive,feng2025embodied,li2026hi}.

\paragraph{Active supervision and modular adaptation.}
Mixture-of-experts, adapters, and LoRA isolate task- or skill-specific
capacity behind a shared backbone~\citep{hu2021lora}.
Robot-learning systems try to construct skill libraries or to route, expand, and update experts as tasks arrive~\citep{luo2026coral,kuzmenko2025moira,zhang2026atomicvlaunlockingpotentialatomic,li2026ditea,clare2025}.
Complementarily, active and corrective imitation learning decide where
additional supervision is needed~\citep{hgdagger,diffdagger,mu2024adademo,khanal2026diseil}.
Trajectory-progress models provide visual signals for assessing task execution~\citep{liang2026robometer,liu2026prm,lee2026roboreward,tan2025robo}.
\methodname{} connects these directions by using progress-based diagnosis
over imagined execution to identify recurring subtask--expert bottlenecks,
request corresponding demonstrations, and apply the acquired supervision to reusable experts.

\section{Method}
\label{sec:method}

\begin{algorithm}[t]
\caption{One \methodname{} coaching round}
\label{alg:robocoach}
\small
\begin{algorithmic}[1]
\REQUIRE Tasks $\mathcal L$, expert library $\experts_b^q$,
         demonstration budget $B_q$, target count $M$
\REQUIRE Frozen $\wm_\theta$, router $\router$, progress judge $J_\phi$,
         completion thresholds, and time limits
\FOR{each task $\ell$ and initial-state trial $i$}
    \FOR{each world-model seed $r\in\{1,2,3\}$}
        \STATE Initialize the imagined trial and route the first subtask
               $g_k$ to expert $e_k$.
        \WHILE{the task remains active}
            \STATE Query the active expert and generate one complete
                   \modelname{} chunk.
            \STATE Commit the generated observations and query the progress judge.
            \IF{the active subtask reaches its completion threshold}
                \STATE Append it to the completed prefix and route the next
                       subtask, or record $\mathsf{SUCCESS}$.
            \ELSIF{the active subtask reaches its time limit}
                \STATE Record $\mathsf{TIMEOUT}(g_k,e_k)$ and terminate.
            \ENDIF
        \ENDWHILE
    \ENDFOR
    \STATE Retain the modal terminal diagnosis if at least two seeds agree.
\ENDFOR
\STATE Rank subtask--expert pairs by task-balanced first-timeout mass
       (Eq.~\ref{eq:first_timeout_mass}) and select the top $M$ pairs.
\STATE Request $B_q/M$ accepted subtask demonstrations for each selected pair.
\STATE Update selected experts using the new demonstrations
       together with old data.
\ENSURE Updated expert library $\experts_b^{q+1}$
\end{algorithmic}
\end{algorithm}

\subsection{Overview}
\label{sec:overview}
\methodname{} uses world-model-imagined execution to decide which subtasks require additional demonstrations and which skill experts should learn from them.
Rather than fitting every complete task with end-to-end demonstrations, we decompose long-horizon execution into reusable atomic skills, each handled by a corresponding expert.
A long-horizon failure can therefore be localized to an unresolved subtask, allowing additional supervision to be collected at the skill level.

Let $q$ index coaching rounds, $\ell\in\mathcal L$ denote a task instruction, and $b$ denote a robot embodiment. Each embodiment uses a shared VLA backbone $\backbone$ together with a library of skill experts
\begin{equation}
    \experts_b^q
    =
    \left\{
        \pol_{b,e}^q
        =
        \backbone \oplus \Delta_{b,e}^q
    \right\}_{e\in\mathcal E_b},
    \label{eq:expert_library}
\end{equation}
where $\mathcal E_b$ indexes the available experts, $\Delta_{b,e}^q$ contains expert $e$'s LoRA parameters~\citep{hu2021lora}, and $\oplus$ denotes applying the expert-specific adapter to the shared
backbone.
The same expert can serve multiple task-specific subtasks, i.e.
\emph{press the red button} and \emph{press the green button} can both invoke
a press expert. We therefore use a subtask--expert pair $(g,e)$ as the
basic unit of coaching.

As illustrated in Fig.~\ref{fig:teaser}, \methodname{} organizes coaching as a
\textbf{Route--Imagine--Diagnose--Improve (RIDI)} loop.
\textbf{Route} selects the next unfinished subtask and dispatches its expert.
\textbf{Imagine} rolls out the active expert inside \modelname{}, our shared
action-conditioned world model, in closed loop.
\textbf{Diagnose} uses a progress judge to determine when the current subtask
has completed. If it remains incomplete until its time limit, the active
subtask--expert pair is recorded.
\textbf{Improve} aggregates these records into a coaching scorecard, requests
new demonstrations for the highest-priority pairs, and updates the corresponding experts.
The updated library then enters the next coaching round, closing the RIDI loop. Algorithm~\ref{alg:robocoach} summarizes a coaching round.

\subsection{Route: from a long-horizon instruction to an atomic expert}
\label{sec:routing}

The router translates a long-horizon instruction into a sequence of
subtask-conditioned expert calls.
Let $k$ index the current subtask and $\mathcal S_b$ denote the available
atomic skills. A subtask $g_k$ specifies a skill and its arguments, such as
the manipulated object or target affordance.
At initialization and after each judge-confirmed completion, the router
receives the instruction $\ell$, current visual observation $o_t$, skill
vocabulary $\mathcal S_b$, and completed-subtask prefix
$\widehat{\mathcal D}_{k-1}$:
\begin{equation}
    g_k=\router\!\left(\ell,o_t,\mathcal S_b,\widehat{\mathcal D}_{k-1}\right),
    \qquad e_k=\eta_b(g_k),
    \label{eq:routing}
\end{equation}
where $\eta_b$ maps subtasks to experts and
$\widehat{\mathcal D}_0=\varnothing$.
Completed subtasks are appended to the prefix before the next routing decision.
The router prompt and output schema appear in Appendix~\ref{app:diagnosis}.

\subsection{Imagine: policy-in-the-loop rollout with \modelname{}}
\label{sec:coachworld}

\begin{figure}[t]
\centering
\includegraphics[width=0.90\textwidth]{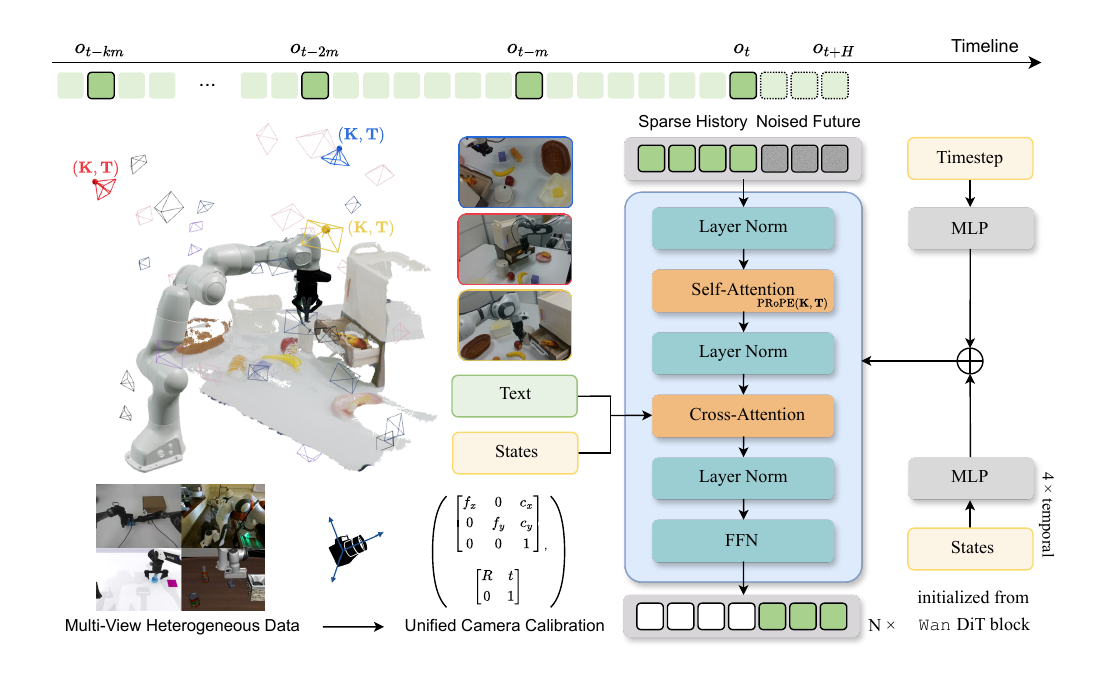}
\caption{\textbf{\modelname{} architecture.}
Sparse visual history, task instructions, and calibrated two-slot
end-effector trajectories condition future-video prediction across
single-arm and bimanual embodiments.}
\label{fig:coachworld_architecture}
\end{figure}

\paragraph{Closed-loop Imagination.}
The active expert rolls in closed loop in \modelname{}. At time $t$, the active expert predicts $a_t=\pol_{b,e_k}^q(o_t,\ell,g_k)$ in delta end-effector space.
A lightweight adapter converts it into a future end-effector trajectory
$\mathbf X_t$, following Ctrl-World~\citep{guo2025ctrl}. For camera $v$, \modelname{} predicts the next video chunk:
\begin{equation}
    \widehat{\mathbf I}_{t+1:t+H}^{\,v}
    \sim \wm_\theta\!\left(
        \mathcal H_t^{\,v},\mathbf X_t,\ell,
        \mathbf K_v,\mathbf T_{v\leftarrow\mathrm{robot}}
    \right),
    \label{eq:wm_rollout}
\end{equation}
where $H$ is the prediction horizon, $\mathcal H_t^{\,v}$ is the sparse
visual history, $\mathbf K_v$ is the camera intrinsic matrix, and
$\mathbf T_{v\leftarrow\mathrm{robot}}$ maps robot coordinates to camera
coordinates.
Each camera stream is predicted independently using its own calibration.
The generated chunk is committed before the next policy or judge query, and its final observation guides the next action chunk, forming a closed-loop rollout.

\paragraph{Shared action and camera conditioning.}
To support heterogeneous robot embodiments within one predictive model,
we represent their actions through a common end-effector interface.
At each resampled trajectory step $n$, arm slot $s\in\{1,2\}$
is represented as $\mathbf x_n^{(s)} = \left[
        \mathbf p_n^{(s)},
        \mathbf r_n^{(s)},
        \gamma_n^{(s)},
        m_n^{(s)} \right],$ where $\mathbf p_n^{(s)}\in\mathbb R^3$ is the end-effector position
in robot coordinates, $\mathbf r_n^{(s)}\in\mathbb R^6$ is its
6D rotation representation, $\gamma_n^{(s)}$ is the gripper state,
and $m_n^{(s)}\in\{0,1\}$ indicates whether the arm slot is present.
Single-arm trajectories populate one slot, whereas bimanual trajectories
populate both.

Camera calibration grounds this trajectory in the visual observation.
The extrinsic transform $\mathbf T_{v\leftarrow\mathrm{robot}}$
first maps end-effector positions into camera coordinates, and
$\mathbf K_v$ projects them into the image plane.
Projected future-state tokens and trajectory rasters then condition
the denoiser on the motion associated with the observed view.
PRoPE~\citep{li2025cameras} further incorporates camera geometry
into visual self-attention.
Together, these components connect the shared 3D action representation
to its view-specific visual consequences (Fig.~\ref{fig:coachworld_architecture}).
Temporal mappings and calibration procedures are detailed in Appendix~\ref{app:data}.

\paragraph{Long-horizon training.}
We train \modelname{} by conditional flow matching~\citep{lipman2022flow} on demonstrations, policy failures, play data, and other non-expert trajectories, using sparse visual history as temporal context.
For a clean future-video latent $z_0$, we sample Gaussian noise
$\epsilon\sim\mathcal N(0,\mathbf I)$ and an interpolation time
$\sigma\sim\mathcal U(0,1)$, and construct
$z_\sigma=(1-\sigma)z_0+\sigma\epsilon$.
The velocity network $v_\theta$ is optimized with
\begin{equation}
    \mathcal L_{\mathrm{CW}}
    =
    \mathbb E
    \left[
        \left\|
            v_\theta(z_\sigma,\sigma\mid c)
            -
            (\epsilon-z_0)
        \right\|_2^2
    \right],
    \label{eq:flow_matching}
\end{equation}
where $c$ contains the sparse visual history, task instruction,
future end-effector trajectory, and camera calibration.
Training details are provided in Appendix~\ref{app:data}.

\subsection{Diagnose: progress-based switching and first-timeout attribution}
\label{sec:diagnosis}

After each generated chunk, a progress judge estimates subtask completion as $p_{k,t}=J_\phi(\mathcal H_t^J,\ell,g_k)\in[0,1]$,
where $\mathcal H_t^J$ is the observation window available up to $t$.
For each subtask--embodiment pair, we calibrate a completion threshold
$\kappa_{g,b}$ and time limit $T^{\max}_{g,b}$ and freeze them before coaching.
If $p_{k,t}\geq\kappa_{g_k,b}$, the subtask is marked complete and control returns to the router. Otherwise, the expert continues until the time limit, measured from its dispatch.
During physical deployment, the same judge and switching procedure use real camera observations.

For trial $i$ of task $\ell$ under world-model seed $r$, we record
\begin{equation}
    Z_{\ell i}^{(r)}=
    \begin{cases}
        \mathsf{SUCCESS}, & \text{if the complete task finishes},\\[2pt]
        \mathsf{TIMEOUT}(g_k,e_k), & \text{if the active subtask times out}.
    \end{cases}
    \label{eq:terminal_diagnosis}
\end{equation}
A timeout terminates the trial and identifies the first unresolved
subtask--expert pair along the executed route.
Each initial-state trial uses three episode-level world-model seeds.
We retain the majority diagnosis $Z_{\ell i}$ only when at least two
seeds agree on the outcome, including the subtask--expert pair for a timeout.
Trials without agreement are excluded from the scorecard.
Judge calibration and evaluation are detailed in Appendix~\ref{app:diagnosis}.

\subsection{Improve: scorecard-guided data acquisition}
\label{sec:allocation}

Let $\mathcal I_\ell^q$ denote the accepted imagined trials for task $\ell$
in round $q$, and $\mathcal L_q^+$ the tasks retained for scorecard
computation. For candidate pair $h=(g,e)$, its task-balanced
first-timeout mass is
\begin{equation}
    \widehat s_b^q(h)
    =\frac{1}{|\mathcal L_q^+|}
    \sum_{\ell\in\mathcal L_q^+}
    \frac{\sum_{i\in\mathcal I_\ell^q}
    \mathbf 1[Z_{\ell i}=\mathsf{TIMEOUT}(h)]}
    {|\mathcal I_\ell^q|}.
    \label{eq:first_timeout_mass}
\end{equation}
The inner term measures how often a task first times out at $h$, while
the outer average gives each task equal weight.
The scorecard also displays imagined success rates, progress traces, and representative failed rollouts. We select the top $M$ pairs and request $B_q/M$ accepted subtask
demonstrations for each. In our experiments, we make $M=2$.
Each request specifies the behavior, object or affordance, and demonstration count, accompanied by imagined failures.
Human operators check safety and collect the requested trajectories.

For each selected expert, new demonstrations are mixed with a smaller
replay sample from existing data for LoRA fine-tuning. Demonstrations from subtasks sharing an expert are pooled for the same update. The backbone and unselected adapters remain unchanged.
The updated library $\experts_b^{q+1}$ then generates the next round's scorecard inside \modelname{}.

\section{Experiments}
\label{sec:experiments}

\begin{table*}[t]
\caption{\textbf{Paired full-episode action-conditioned prediction.}
All models receive matched visual histories and future actions on DROID-180 and
Lab Franka-180.
The two \modelname{} variants compare mixed- and single-domain training at
matched training volume.
LPIPS/FVD measure video fidelity, HSD/nDTW/DYN measure end-effector trajectory
agreement, and Phys.\ Adh./Instr.\ Follow.\ are mean blind human ratings on a 1--5 scale, divided by five.
Best and second-best values within each evaluation set are bold and underlined,
respectively.}
\label{tab:wm_compact}
\centering
\scriptsize
\setlength{\tabcolsep}{3.7pt}
\renewcommand{\arraystretch}{1.00}
\resizebox{\textwidth}{!}{%
\begin{tabular}{llccccccc}
\toprule
Method & Evaluation set & LPIPS$\downarrow$ & FVD$\downarrow$ &
HSD$\uparrow$ & nDTW$\uparrow$ & DYN$\uparrow$ &
Phys. Adh.$\uparrow$ & Instr. Follow.$\uparrow$ \\
\midrule

Ctrl-World
& DROID-180
& 0.2078 & 102.97 & 0.2234 & 0.2323 & 0.0668 & 0.6067 & 0.7867 \\
& Lab Franka-180
& 0.3060 & 493.20 & 0.1304 & 0.1037 & 0.0458 & 0.2600 & 0.2400 \\

Cosmos~3
& DROID-180
& 0.2830 & 140.62 & 0.1702 & 0.1698 & 0.0571 & 0.7267 & 0.5467 \\
& Lab Franka-180
& 0.3397 & 634.75 & 0.1658 & 0.1374 & 0.0653 & \underline{0.3900} & 0.2467 \\

OSCAR-2B
& DROID-180
& 0.2248 & 101.92 & 0.1785 & 0.1838 & 0.0456 & 0.5267 & 0.5467 \\
& Lab Franka-180
& 0.2014
& \textbf{236.78}
& \textbf{0.1909}
& \textbf{0.2110}
& \textbf{0.1412}
& \textbf{0.4800}
& \textbf{0.5833} \\

\midrule

\modelname{} (mixed-domain)
& DROID-180
& \textbf{0.0996}
& \textbf{51.71}
& \textbf{0.2633}
& \textbf{0.2812}
& \textbf{0.0959}
& \underline{0.7467}
& \textbf{0.8800} \\
& Lab Franka-180
& \textbf{0.1712}
& \underline{284.76}
& 0.1770
& \underline{0.1767}
& 0.1113
& 0.3333
& \underline{0.5333} \\

\modelname{} (single-domain)
& DROID-180
& \underline{0.1067}
& \underline{56.56}
& \underline{0.2441}
& \underline{0.2599}
& \underline{0.0723}
& \textbf{0.7533}
& \underline{0.8667} \\
& Lab Franka-180
& \underline{0.1768}
& 297.56
& \underline{0.1854}
& 0.1754
& \underline{0.1137}
& 0.3000
& 0.4267 \\

\bottomrule
\end{tabular}
}
\end{table*}

\subsection{Evaluation setting}

We evaluate \methodname{} on ten LIBERO-Long tasks and five RoboTwin~2.0 tasks (following Lingbot-VA~\citep{li2026causal}'s selection), together with real-robot tasks on Franka Research 3 and AgileX dual-Piper.
We use $\pi_{0.5}$~\citep{intelligence2025pi_} as the shared VLA backbone
for LIBERO and Franka, and MolmoAct2~\citep{fang2026molmoact2actionreasoningmodels}
for RoboTwin and AgileX.
Each task is evaluated over 50 trials in simulation or 20 trials on real robots.
Within each domain, coaching comparisons match demonstration budgets,
LoRA training settings, and evaluation protocols. Appendix~\ref{app:protocols}
describes the tasks and data splits, and Appendix~\ref{app:coaching} details
the coaching protocol.

We first assess \modelname{} prediction fidelity, agreement with deployed policy outcomes,
and progress-judge reliability. We then measure complete-task improvement under a fixed demonstration budget and separate data acquisition from update location. Finally, we test whether the coached skill experts can generalize to held-out compositions.

\subsection{Evidence for world-model-guided intervention}
\label{sec:exp_evidence}

Using \modelname{} as an active coach requires three properties. First, its predicted
futures must be faithful to the observed interaction. Second, policy rollouts inside the model must preserve the success profile observed in the deployed
environment. Third, the progress judge must reliably detect completion and attribute failures to specific bottlenecks. We evaluate these requirements in the same order: full-episode prediction, policy-level factuality, and progress switching with failure attribution.

\textbf{World-model fidelity on held-out episodes.}
We first evaluate \modelname{} as a full-episode simulator on held-out test sets. We construct two test sets by selecting 180 video clips respectively from the DROID dataset and from our self-collected dataset. Each model receives identical inputs.
We use PSNR, SSIM, LPIPS, and FVD to measure appearance fidelity.
We use HSD, nDTW, and DYN to measure end-effector trajectory and action-following fidelity, following EWMBENCH~\citep{yue2025ewmbenchevaluatingscenemotion}. We evaluate
Physics Adherence and Instruction Following with $1$--$5$ rubrics adapted from MiraBench~\citep{yang2026mirabenchevaluatingactionconditionedreliability} and WorldArena~\citep{shang2026worldarena}.
The two \modelname{} variants compare mixed- and single-domain training at matched training volume.

Table~\ref{tab:wm_compact} compares full-episode predictions.
On DROID-180, mixed-domain \modelname{} achieves the best LPIPS ($0.0996$),
FVD ($51.71$), all 3 trajectory metrics, and instruction-following score
($0.880$). On Lab Franka-180, it achieves the best LPIPS ($0.1712$). Mixed-domain training improves most visual and trajectory measures, indicating that greater domain diversity makes world models more powerful at future prediction. Appendix~\ref{app:wm_results} gives
supplemental results and evaluation details.

\begin{wrapfigure}{r}{0.49\textwidth}
    \centering
    \includegraphics[width=\linewidth]
    {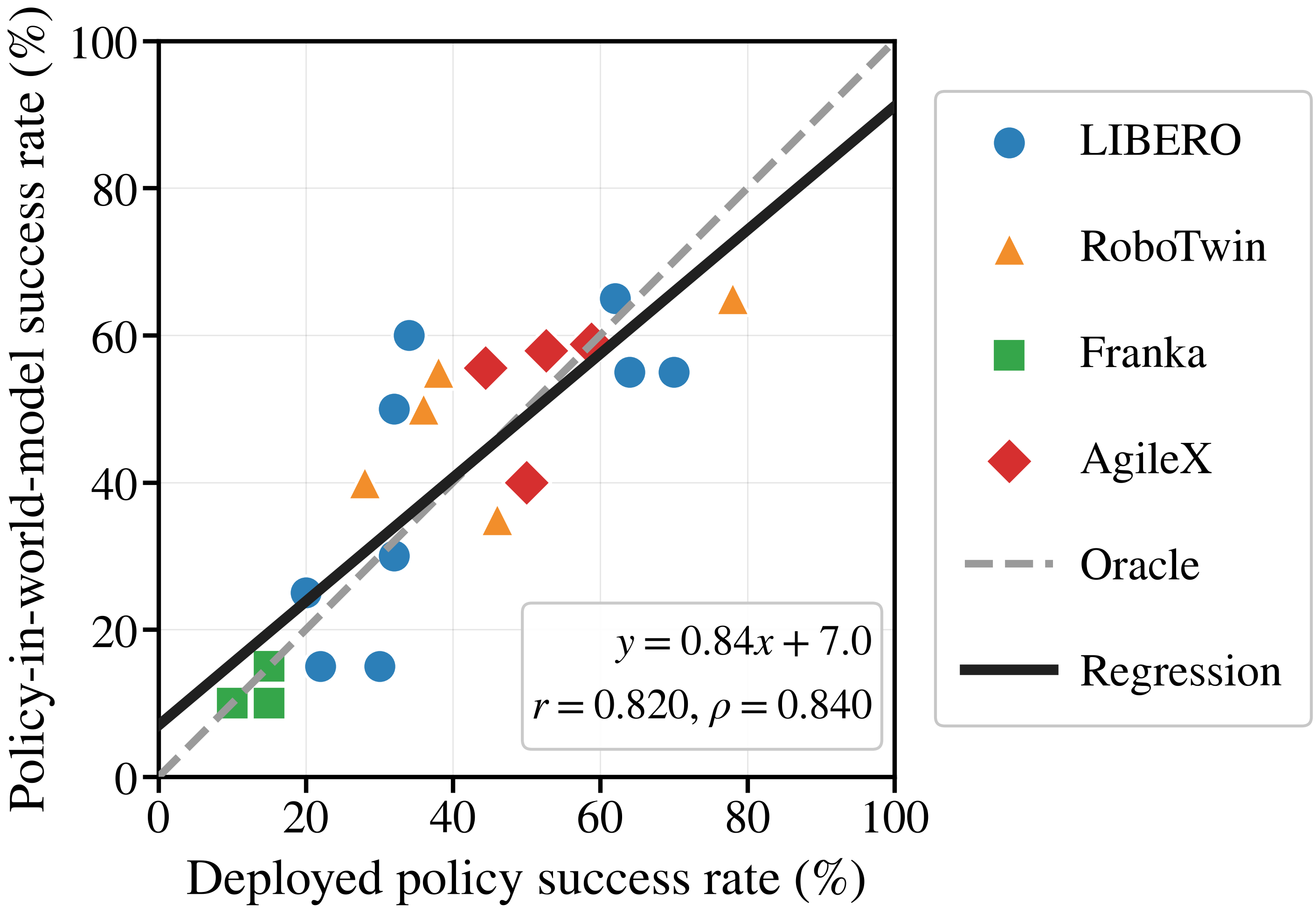}
    \caption{\textbf{Imagined versus deployed policy success.} All four platforms share one coordinate system and one pooled OLS fit.}
    \label{fig:policy_success_correlation}
\end{wrapfigure}

\textbf{Agreement with deployed policy outcomes.}
Useful simulators must reflect policy failures as well as successes. We thereby compare complete-task success under policy-in-the-loop imagination with matched deployed executions. This study uses 22 frozen task--policy
checkpoint pairs, one per task, with matched initial-state strata and trial counts. Imagined and deployed success have a Pearson correlation of
$r=0.820$ and a Spearman correlation of $\rho=0.840$
(Fig.~\ref{fig:policy_success_correlation}), showing agreement in task-level
performance differences across the evaluated pairs.

\textbf{Progress-judge reliability and failure attribution.}
The progress judge is a core component of the active-coaching loop.
During execution, it determines whether the active subtask is complete and
whether control should switch to the next expert.
An incorrect switch can corrupt the remaining route, while an incorrect timeout
can corrupt the resulting failure attribution.
To isolate judge quality from world-model generation, we provide every candidate judge with the same causal reference-video replay.
We evaluate three aspects of judge reliability.
Switch MAE measures temporal error on subtask-transition boundaries,
while Recall@1.0s measures the fraction of ground-truth transitions recovered
within one second.
Spearman $\rho$ measures rank agreement between predicted and ground-truth
terminal stages over task--stage cells.
We compute $\rho$ separately on LIBERO, RoboTwin, real single-arm, and real dual-arm settings and report their unweighted macro average.
Outcome precision, recall, and F1 evaluate binary transition decisions on adjacent-stage examples.

We compare Qwen3-VL~\citep{bai2025qwen3vl}, LIV~\citep{ma2023liv},
Contrastive $\lambda$~\citep{goko2025contrastive},
SuccessVQA~\citep{du2023successvqa}, TOPReward~\citep{chen2026topreward},
and RoboMeter~\citep{liang2026robometer}.
Appendix~\ref{app:judge_baselines} describes their implementations.
RoboMeter achieves the best scores under this protocol
(Appendix Table~\ref{tab:diagnosis}), including a Switch MAE of $0.414$\,s,
$82.73\%$ Recall@1.0s, and $88.21\%$ outcome F1.
We use it as the progress judge in \methodname{}.

\begin{table*}[!htbp]
\caption{\textbf{RoboMeter reliability under reference and world-model observations.}
We test RoboMeter on reference video and \modelname{}-generated rollout.
Switch MAE and Recall@1.0s measure stage-transition timing, Spearman $\rho$
measures terminal-stage rank agreement, and outcome precision, recall, and F1
measure binary transition
decisions on adjacent-stage examples.}
\label{tab:robometer_judge}

\centering
\scriptsize
\setlength{\tabcolsep}{3.5pt}
\renewcommand{\arraystretch}{1.05}

\resizebox{\textwidth}{!}{%
\begin{tabular}{lccccccc}
\toprule
Observation source
& Switch MAE (s)$\downarrow$
& Switch Rec.\@1.0s (\%)$\uparrow$
& Spearman $\rho$$\uparrow$
& TP/TN/FP/FN
& Out.-Prec. (\%)$\uparrow$
& Out.-Rec. (\%)$\uparrow$
& Out.-F1 (\%)$\uparrow$ \\
\midrule

Reference video
& 0.414
& 82.73
& 0.727
& 101/92/18/9
& 84.87
& 91.82
& 88.21 \\

\modelname{} rollout
& 0.815
& 62.73
& 0.638
& 96/87/23/14
& 80.67
& 87.27
& 83.84 \\

\bottomrule
\end{tabular}%
}

\end{table*}

We further evaluate RoboMeter on \modelname{}-generated observations under the
same diagnostic protocol (Table~\ref{tab:robometer_judge}).
Under generated observations, RoboMeter achieves a terminal-stage Spearman
correlation of $\rho=0.638$, an outcome F1 of $83.84\%$, a Switch MAE of
$0.815$\,s, and $62.73\%$ Recall@1.0s.
These results show that the diagnostic signals required for active coaching
remain informative under world-model-generated observations.

\subsection{Fixed-budget coaching and intervention coupling}
\label{sec:exp_coaching}

\begin{figure*}[h]
    \centering
    \includegraphics[width=\textwidth]{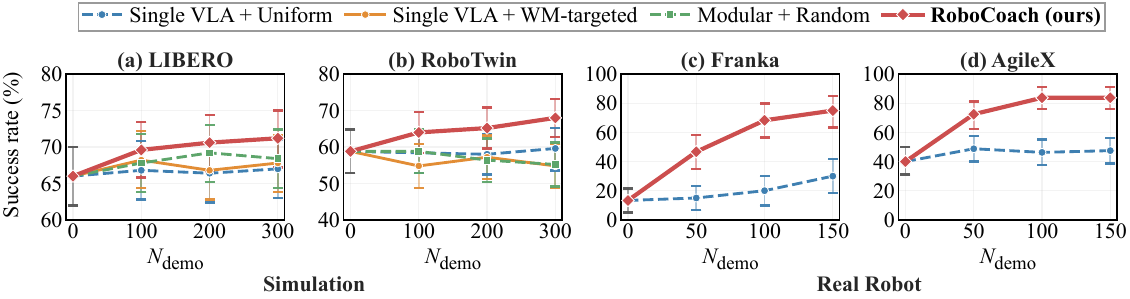}
    \caption{\textbf{Complete-task success over coaching rounds.}
We evaluate task-averaged success rate as cumulative coaching demonstrations increase.
Error bars show 95\% confidence intervals for task-averaged success,
computed from per-task evaluation counts (Appendix~\ref{app:coaching}). As a result, \methodname{} achieves the highest success rate in all four settings.}
    \label{fig:coaching_curve}
\end{figure*}

We next test whether coaching improves complete-task success and which
acquisition and update choices contribute to the improvement.
Each coaching round adds $B_{\mathrm{sim}}=100$ accepted subtask demonstrations
in simulation or $B_{\mathrm{real}}=50$ on each real-robot platform.
We report complete-task success averaged equally across tasks at cumulative
budgets $\{0,B,2B,3B\}$, together with normalized Budget AUC.
We evaluate all four conditions in Table~\ref{tab:coaching_controls} in
simulation and compare \emph{Single VLA + Uniform} with \methodname{} on
real robots due to hardware effort.

\begin{table}[h]
\centering
\caption{\textbf{Acquisition and update conditions.}}
\label{tab:coaching_controls}
\small
\setlength{\tabcolsep}{6pt}
\begin{tabular}{@{}lll@{}}
\toprule
Condition & Data selection & Update location \\
\midrule
Single VLA + Uniform & Uniform & Shared global adapter \\
Single VLA + WM-targeted & WM-targeted & Shared global adapter \\
Modular + Random & Random subtask--expert pairs & Selected skill experts \\
\methodname{} & WM-targeted & Selected skill experts \\
\bottomrule
\end{tabular}
\end{table}

All conditions freeze the VLA backbone and only train LoRA adapters.
\emph{Single VLA + Uniform} acquires demonstrations uniformly across all tasks and updates one shared adapter. \emph{Single VLA + WM-targeted} receives the
same amount of substask demonstrations acquired for \methodname{}, but applies a shared
update rather than updating the corresponding skill experts.
\emph{Modular + Random} selects two subtask--expert pairs at random and
updates their associated experts, pooling demonstrations when targets share an expert. The comparisons test targeted acquisition
under shared adaptation, shared versus expert-specific updates on identical acquired data, and target selection within the modular system.

Figure~\ref{fig:coaching_curve} shows complete-task success over three coaching
rounds.
On real robots, both methods start at $13.3\%$ on Franka and $40.0\%$ on
AgileX.
After $150$ additional subtask demonstrations per platform, \methodname{} reaches $75.0\%$
and $83.8\%$, compared with $30.0\%$ and $47.5\%$ for
\emph{Single VLA + Uniform}.
Normalized Budget AUC is $53.1$ versus $18.9$ on Franka and $72.7$ versus $46.3$ on AgileX.
In simulation, \methodname{} reaches $71.2\%$ on LIBERO and $68.0\%$ on RoboTwin, the highest final success among the four compared conditions.

With a shared adapter, targeted acquisition changes final success relative
to uniform acquisition by $+0.8$ pp on LIBERO and $-4.8$ pp on RoboTwin.
Applying the same targeted demonstrations to selected experts instead adds
$3.4$ and $13.2$ pp.
Within the modular system, world-model targeting exceeds random target selection by $2.8$ and $12.8$ pp.
The strongest results come from combining targeted demonstrations with updates
to their corresponding skill experts.
Appendix~\ref{app:coaching} reports per-round results and acquisition traces.

\subsection{Held-out composition generalization}
\label{sec:exp_reuse}

\begin{figure*}[!htbp]
    \centering
    \includegraphics[width=\textwidth]{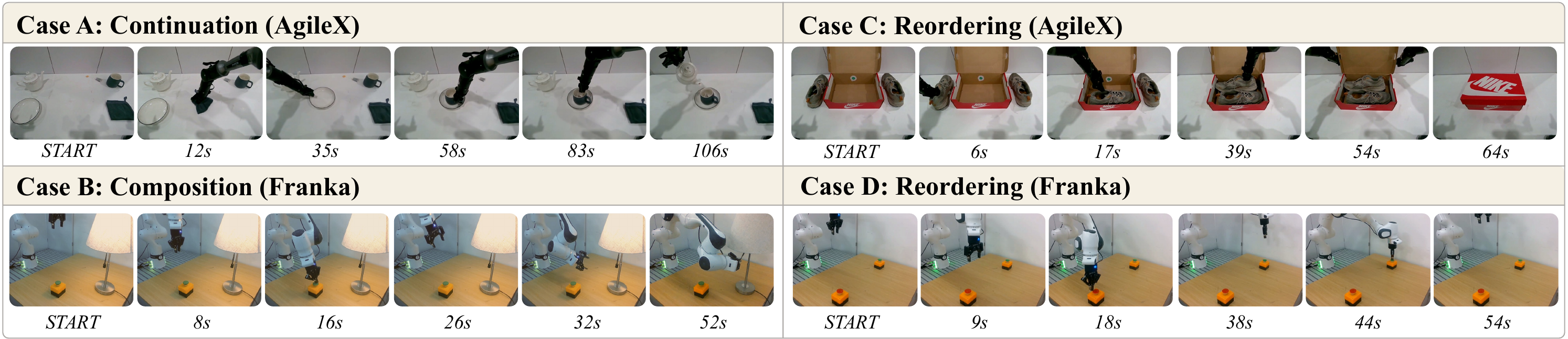}
    \caption{\textbf{Held-out composition case studies.}
    Execution sequences on AgileX (A, C) and Franka (B, D),
    illustrating longer continuation (A), cross-task composition (B),
    and skill reordering (C, D).
    Frames in each sequence progress from left to right.}
    \label{fig:4case22}
\end{figure*}

Finally, we test whether coached experts can be recombined into four
long-horizon task routes excluded from training and coaching, with 20
real-robot trials per route. Figure~\ref{fig:4case22} shows example executions,
and Appendix~\ref{app:composition} gives the route definitions and protocol.

\methodname{} succeeds in $13/20$ trials ($65\%$) on the AgileX
continuation combining table setting with tea making (Case A), and in
$3/20$ trials ($15\%$) when combining green-button pressing with lamp-cord
pulling on Franka (Case B). For skill reordering, it succeeds in $5/20$
trials ($25\%$) when reversing shoe placement before closing the box on
AgileX (Case C), and in $7/20$ trials ($35\%$) when reversing the button
order on Franka using the same press expert (Case D).
\emph{Single VLA + Uniform} succeeds in none of the four cases.
The four-route average is $35.0\%$ for \methodname{} and $0\%$ for the
baseline. These results show that the coached experts can be recombined along unseen routes.

\section{Conclusion}

\methodname{} uses imagined execution to decide which demonstrations to
collect and which reusable skill experts should learn from them.
Its RIDI loop connects world-model prediction, progress-based diagnosis,
and targeted supervision to iterative policy improvement.
Experiments show that applying the same acquired demonstrations to
corresponding experts outperforms a shared update.
The resulting experts greatly improve real-robot success under fixed demonstration budgets and remain reusable on previously unseen compositions for generalization.
Together, these results support a role for world models as active coaches that guide what a robot should learn next.

\paragraph{Limitations.}
When the manipulated object is occluded, or when contact and collision occur outside the camera's view, physical modeling and prediction in that region become difficult. \methodname{} likewise relies on action-conditioned predictions that preserve task-relevant evidence, which remains challenging under occlusion and out-of-view interaction.
Agreement across generation seeds reduces sensitivity to stochastic
variation but cannot rule out systematic world-model bias.
The expert library is predefined by skill semantics. Learning its structure
and reducing the need for human demonstrations remain open directions.

\section*{Acknowledgments}
We sincerely thank Yanjiang Guo for his help throughout this project.

\bibliography{references}
\bibliographystyle{plainnat}

\clearpage
\beginappendix
\section{Experimental Assets and Provenance}
\label{app:protocols}

\paragraph{Splits and evaluation units.}
All views from one physical episode share the same split.
DROID-180 contains 180 view-specific evaluation episodes from 90 physical episode groups with two static views each; laboratory Franka-180 contains 180 human-reviewed trajectories from three tasks and is fully disjoint from world-model training.
The DROID manifest is frozen before any model is scored and is not selected using per-model metrics.
Judge-training trajectories, coaching demonstrations, and final policy-evaluation episodes remain disjoint at the physical-episode level.
The held-out composition sequences and their semantic paraphrases are excluded from policy and judge training, coaching data, and router examples.
Final LIBERO, RoboTwin, Franka, and AgileX evaluation counts are 50, 50, 20 and 20 respectively.

\noindent\textbf{Compute resources.}
The main model-training and offline evaluation experiments were conducted
using eight NVIDIA H100 GPUs.

\subsection{Real-Robot Platforms and Tasks}
\label{app:real_setup}

\begin{figure}[!ht]
\centering
\begin{minipage}[t]{0.485\linewidth}
    \centering
    \includegraphics[width=\linewidth]{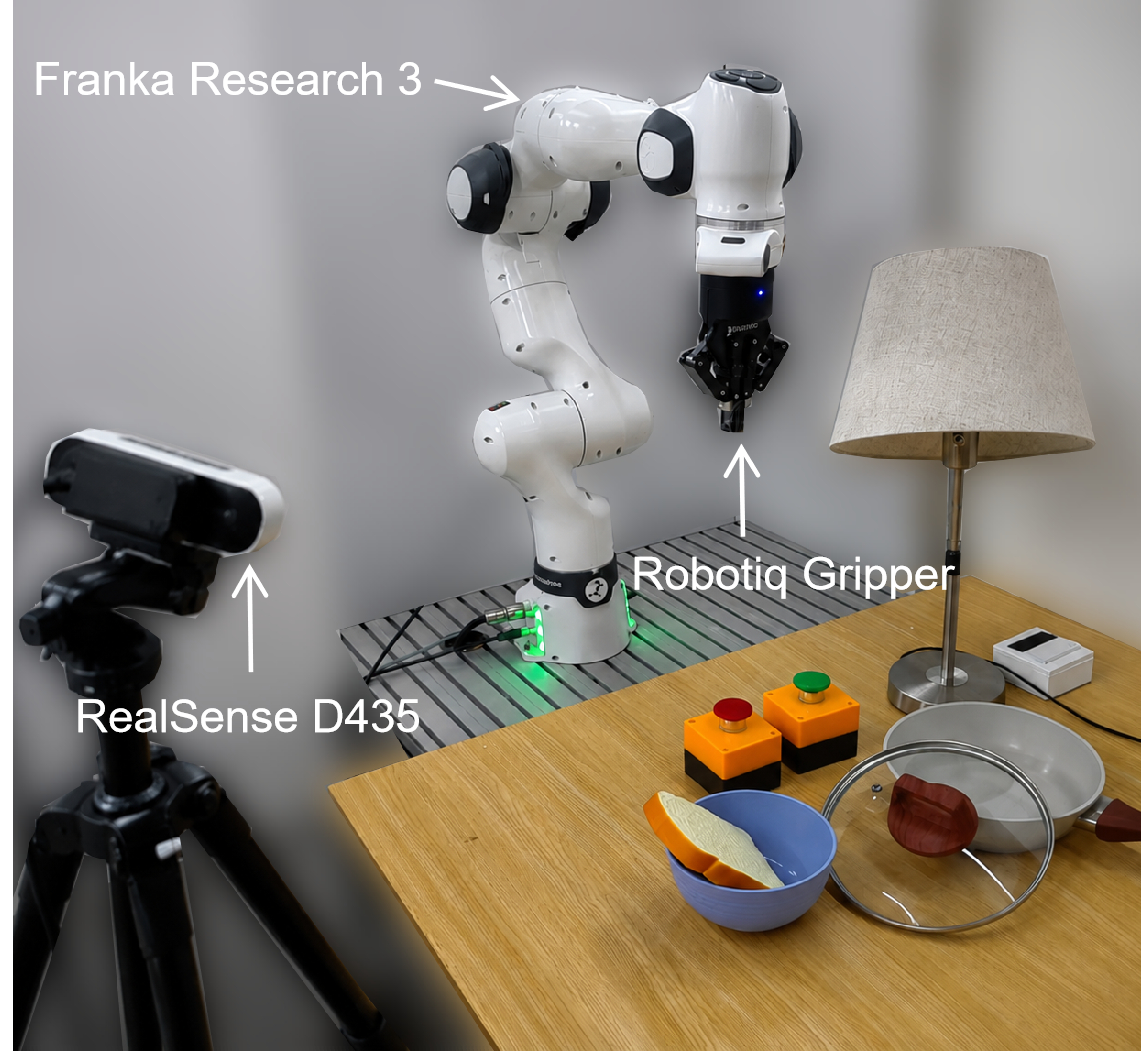}\\[-0.3em]
    \textbf{(a) Franka Research 3 with a Robotiq gripper}
\end{minipage}
\hfill
\begin{minipage}[t]{0.485\linewidth}
    \centering
    \includegraphics[width=\linewidth]{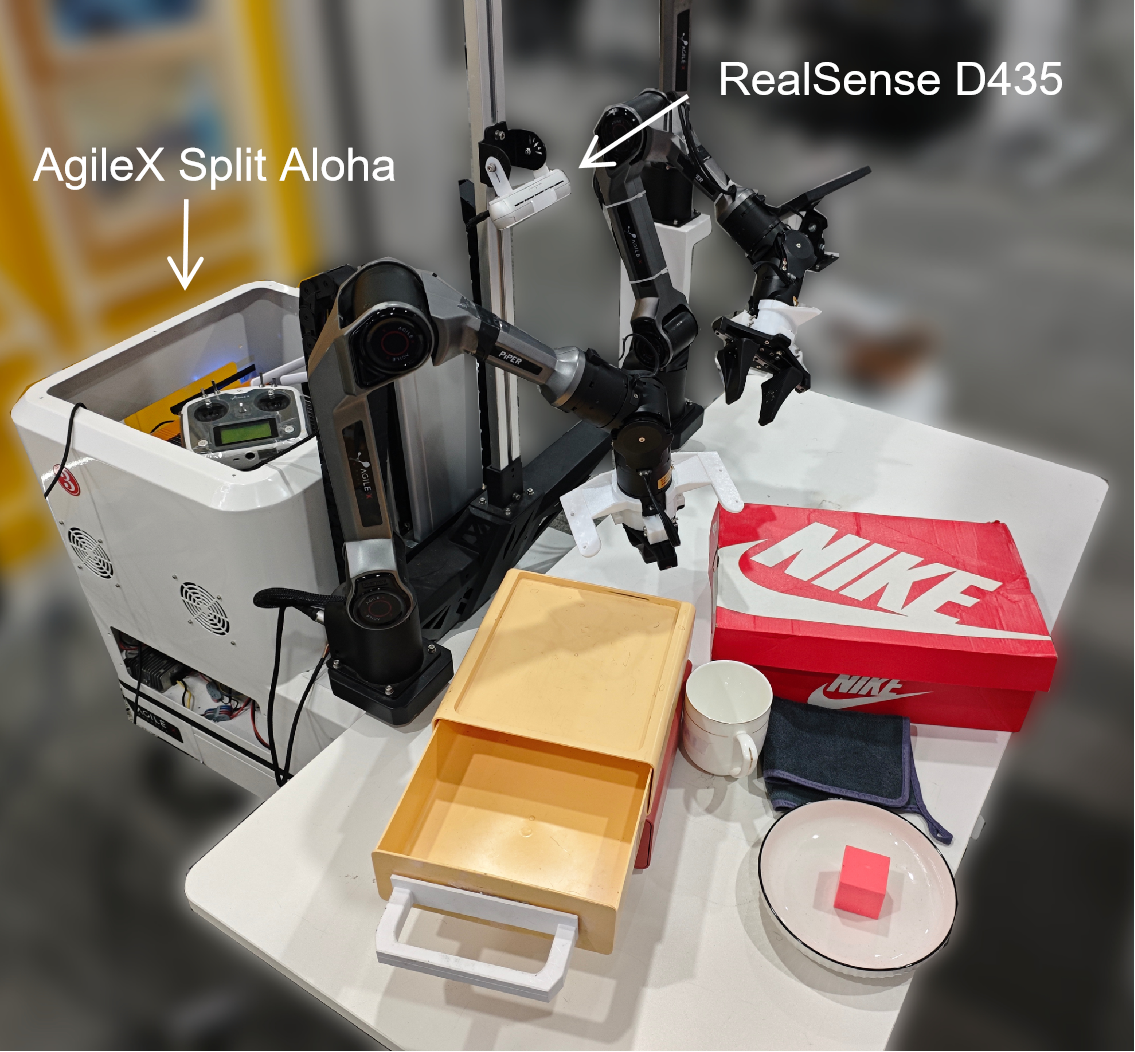}\\[-0.3em]
    \textbf{(b) AgileX Split Aloha with dual Piper arms}
\end{minipage}
\caption{\textbf{Real-robot platforms.}}
\label{fig:real_setups}
\end{figure}

\begin{table*}[tb]
\caption{\textbf{Real-robot task corpus and ordered decompositions.}}
\label{tab:real_tasks}
\centering
\footnotesize
\setlength{\tabcolsep}{4pt}
\begin{tabular}{llp{10.5cm}}
\toprule
Platform & Task & Ordered subtasks \\
\midrule

Franka
& Lamp switch and cord
& press the lamp switch; pull the lamp cord \\

Franka
& Two buttons press
& press the green button; press the red button \\

Franka
& Bread and pan
& pick the bread; place the bread into the pan; pick the lid; place the lid over the pan \\

\midrule

AgileX
& Block and drawer
& pick the red block; place the block into the drawer; close the drawer \\

AgileX
& Shoes and box
& pick the right shoe; place the right shoe into the shoebox;
pick the left shoe; place the left shoe into the shoebox; close the box \\

AgileX
& Table setting
& pick the cloth; wipe the table; push the plate to the center of the table;
pick the cup; place the cup on the plate \\

AgileX
& Tea making
& pick up the tea bag; place the tea bag into cup; pour water into the cup \\

\bottomrule
\end{tabular}
\end{table*}

\begin{table*}[t]
\caption{\textbf{Length distribution of the current full-episode world-model suites.}
Duration is measured on the 5\,Hz evaluation timeline after the initial history.}
\label{tab:full_episode_lengths}
\centering
\scriptsize
\setlength{\tabcolsep}{4pt}
\begin{tabular}{lrrrrrr}
\toprule
Evaluation set & Episodes & Min (s) & Q1 (s) & Median (s) & Q3 (s) & Max (s) \\
\midrule
DROID-180 & 180 & 5.8 & 10.2 & 14.2 & 23.9 & 95.4 \\
Laboratory Franka-180 & 180 & 17.0 & 33.8 & 37.6 & 44.3 & 66.8 \\
\bottomrule
\end{tabular}
\end{table*}

\subsection{Simulation Task Suites}
\label{app:simulation_tasks}

LIBERO-Long provides ten single-arm long-horizon tasks, while RoboTwin~2.0 provides the five dual-arm horizon-3 tasks according to Lingbot-VA used in our controlled study~\citep{liu2023libero,robotwin2,li2026causal}.
The following counts describe the available demonstration pools; evaluation uses disjoint simulator seeds and initial states.

\begin{table}[!ht]
\caption{\textbf{LIBERO-Long task pool.}
Task IDs L0--L9 are used throughout the per-task coaching results in the appendix.
Task names are shortened versions of the original language instructions.
The pool contains 379 successful demonstrations.}
\label{tab:libero_long_tasks}
\centering
\scriptsize
\setlength{\tabcolsep}{3.2pt}
\begin{tabular}{clrr@{\hspace{1.2em}}clrr}
\toprule
ID & Task & Demos & Avg. steps
& ID & Task & Demos & Avg. steps \\
\midrule
L0 & Soup + tomato sauce in basket
& 38 & 258.1
& L5 & Book in rear caddy compartment
& 33 & 290.0 \\

L1 & Cream cheese + butter in basket
& 36 & 250.6
& L6 & White mug on plate + pudding right
& 29 & 407.2 \\

L2 & Turn on stove + place moka pot
& 34 & 293.8
& L7 & Soup + cream cheese in basket
& 49 & 259.2 \\

L3 & Black bowl in drawer + close
& 41 & 265.0
& L8 & Both moka pots on stove
& 35 & 245.1 \\

L4 & Two mugs on left/right plates
& 43 & 267.3
& L9 & Mug in microwave + close
& 41 & 186.2 \\
\bottomrule
\end{tabular}
\end{table}

\begin{table}[!ht]
\caption{\textbf{RoboTwin~2.0 task pool.}
We use five horizon-3 tasks in the controlled coaching study.
Task IDs R0--R4 are used throughout the per-task coaching results in the appendix.
Each task contains 50 clean and 500 randomized demonstrations, for 2,750 trajectories in total.}
\label{tab:robotwin_tasks}
\centering
\footnotesize
\setlength{\tabcolsep}{4pt}
\begin{tabular}{clrp{3.0in}}
\toprule
ID & Task & Avg. steps & Instruction \\
\midrule
R0 & Blocks Ranking RGB
& 466
& Arrange the red, green, and blue blocks from left to right. \\

R1 & Blocks Ranking Size
& 466
& Arrange the three blocks from largest to smallest. \\

R2 & Put Bottles Dustbin
& 637
& Place the bottles into the dustbin on the left. \\

R3 & Stack Bowls Three
& 476
& Stack the three bowls. \\

R4 & Stack Blocks Three
& 481
& Stack blue on green and green on red. \\
\bottomrule
\end{tabular}
\end{table}

\FloatBarrier

\section{\modelname{} Training and Closed-Loop Interface}
\label{app:data}

\subsection{Data Mixture and Admission}
Besides self-collected data, \modelname{} uses DROID and RoboMIND~1.0 Franka for real single-arm manipulation, RoboCOIN, ViFailBack, and GigaAI dual-Piper for real bimanual manipulation, and LIBERO and RoboTwin for simulation~\citep{khazatsky2024droid,wu2025robomind,wu2025robocoin,vifailback,liu2023libero,robotwin1,robotwin2}.
The mixture is about 500 hours, including successful demonstrations and sub-optimal data.
Synchronized views form separate single-view samples with shared arm-slot trajectories and view-specific calibration.
We use source or simulator calibration when available and recover missing transforms as described below.
Samples with invalid state, unresolved synchronization, or failed geometric alignment are excluded.
Mixed-domain training uses the complete pool, whereas single-domain training uses target-domain data at matched total training volume.

\begin{table}[t]
\caption{\textbf{\modelname{} training configuration.}}
\label{tab:wm_training_hparams}
\centering
\footnotesize
\renewcommand{\arraystretch}{1.08}
\begin{tabular}{ll}
\toprule
Item & Setting \\
\midrule
Initialization & Wan2.2 TI2V-5B \\
Frequency & 5\,Hz \\
Image size & $512\times768$ \\
History:future latents & 5:3 \\
Optimizer & AdamW with cosine decay \\
Learning rate & $1\times10^{-5}$; $5\times10^{-6}$ late stage \\
Batch & 40 \\
\bottomrule
\end{tabular}
\end{table}

\subsection{Closed-Loop Temporal Contract}
For each embodiment, a frozen timestamp map resamples a policy action chunk onto the canonical EEF clock and aligns it with one decoded video chunk.
The complete EEF trajectory and generated RGB chunk are committed before the next policy or judge query; the terminal commanded EEF state initializes the next action chunk.

\subsection{Camera and Action Grounding}
Each embodiment adapter maps native policy actions to the same two-slot EEF representation.
At every canonical EEF timestamp, a slot stores metric position, 6D rotation, normalized gripper state, and an existence mask; single-arm data populate one slot and bimanual data populate both.
The adapter also records the native policy timestamps used to construct the canonical trajectory, making temporal resampling part of the frozen interface rather than an implicit preprocessing choice.

$T_{v\leftarrow\mathrm{robot}}\in SE(3)$ maps metric positions from the canonical robot frame into camera $v$, and $K_v$ includes all resize, crop, and padding operations.
For $[\tilde p_x,\tilde p_y,\tilde p_z,1]^\top=T_{v\leftarrow\mathrm{robot}}[p^\top,1]^\top$ and $\tilde p_z>0$,
\begin{equation}
    [\bar u,\bar v,1]^\top
    =\tilde p_z^{-1}K_v[\tilde p_x,\tilde p_y,\tilde p_z]^\top.
    \label{eq:camera_projection}
\end{equation}
Points behind the camera or outside the valid image mask do not contribute trajectory-raster tokens.

\paragraph{Camera calibration.}
We adapt the coarse-to-fine procedure of CalibAll~\citep{xie2025unify} when a
camera-to-robot transform is unavailable. A coarse transform is initialized
from robot--image correspondences and refined by aligning differentiably
rendered robot geometry with image-space robot masks. Source intrinsics are
retained when available, with resizing, cropping, and padding folded into
$K_v$. We admit a camera stream only after projected forward-kinematic EEF
traces align with the observed motion. Whereas CalibAll expresses actions in
the camera frame, \modelname{} retains metric EEF states in the canonical robot
frame and uses the recovered transform to construct view-specific trajectory
tokens and rasters.

\subsection{Training Objective}

We use the conditional flow-matching objective in
Eq.~\ref{eq:flow_matching}.
For camera $v$, the conditioning is
\[
c=
\left(
\mathcal H_t^{\,v},
\mathbf X_t,
\ell,
\mathbf K_v,
\mathbf T_{v\leftarrow\mathrm{robot}}
\right),
\]
where $\mathbf X_t$ is the future end-effector trajectory defined in
Sec.~\ref{sec:coachworld}.
\modelname{} uses five sparse-history latents and predicts three future latents.

\FloatBarrier

\section{Additional \modelname{} Evidence}
\label{app:wm_results}

\subsection{Evaluation Package}
Each evaluated stream records its checkpoint, physical episode, camera, initial-history endpoint, action sequence, generation seed, prediction, and metrics.
FVD-16 uses a frozen early/middle/late clip manifest (540 clips per 180-episode domain) and an episode-grouped bootstrap.

\subsection{Appearance Fidelity and Long-Horizon Stability}

\begin{table}[!ht]
\caption{\textbf{Supplemental full-episode visual metrics.}
PSNR and SSIM are secondary appearance diagnostics. The \modelname{} variants
compare mixed-domain and single-domain training at matched volume.}
\label{tab:wm_visual_supplement}
\centering
\footnotesize
\setlength{\tabcolsep}{4pt}
\begin{tabular}{lcccc}
\toprule
& \multicolumn{2}{c}{DROID-180} & \multicolumn{2}{c}{Lab Franka-180} \\
\cmidrule(lr){2-3}\cmidrule(lr){4-5}
Method & PSNR$\uparrow$ & SSIM$\uparrow$ & PSNR$\uparrow$ & SSIM$\uparrow$ \\
\midrule
Ctrl-World & 20.597 & 0.7972 & 21.859 & 0.8279 \\
Cosmos~3 & 15.776 & 0.5916 & 18.308 & 0.6654 \\
OSCAR-2B & 17.304 & 0.7449 & 21.915 & 0.8473 \\
\modelname{} (single-domain) & 21.115 & 0.8696 & \textbf{22.193} & 0.8649 \\
\modelname{} (mixed-domain) & \textbf{21.438} & \textbf{0.8768} & 22.168 & \textbf{0.8665} \\
\bottomrule
\end{tabular}
\end{table}

PSNR, SSIM, and LPIPS exclude the shared conditioning frame, average future frames within each episode, and then weight episodes equally.
The equal-domain macro for single-domain training is $21.654/0.8673/0.1418$ (PSNR/SSIM/LPIPS), compared with $21.803/0.8716/0.1354$ for mixed-domain training. FVD remains domain-specific and is never macro-averaged across the two suites.

\begin{figure*}[t]
\centering
\includegraphics[height=0.4\textheight,keepaspectratio]
{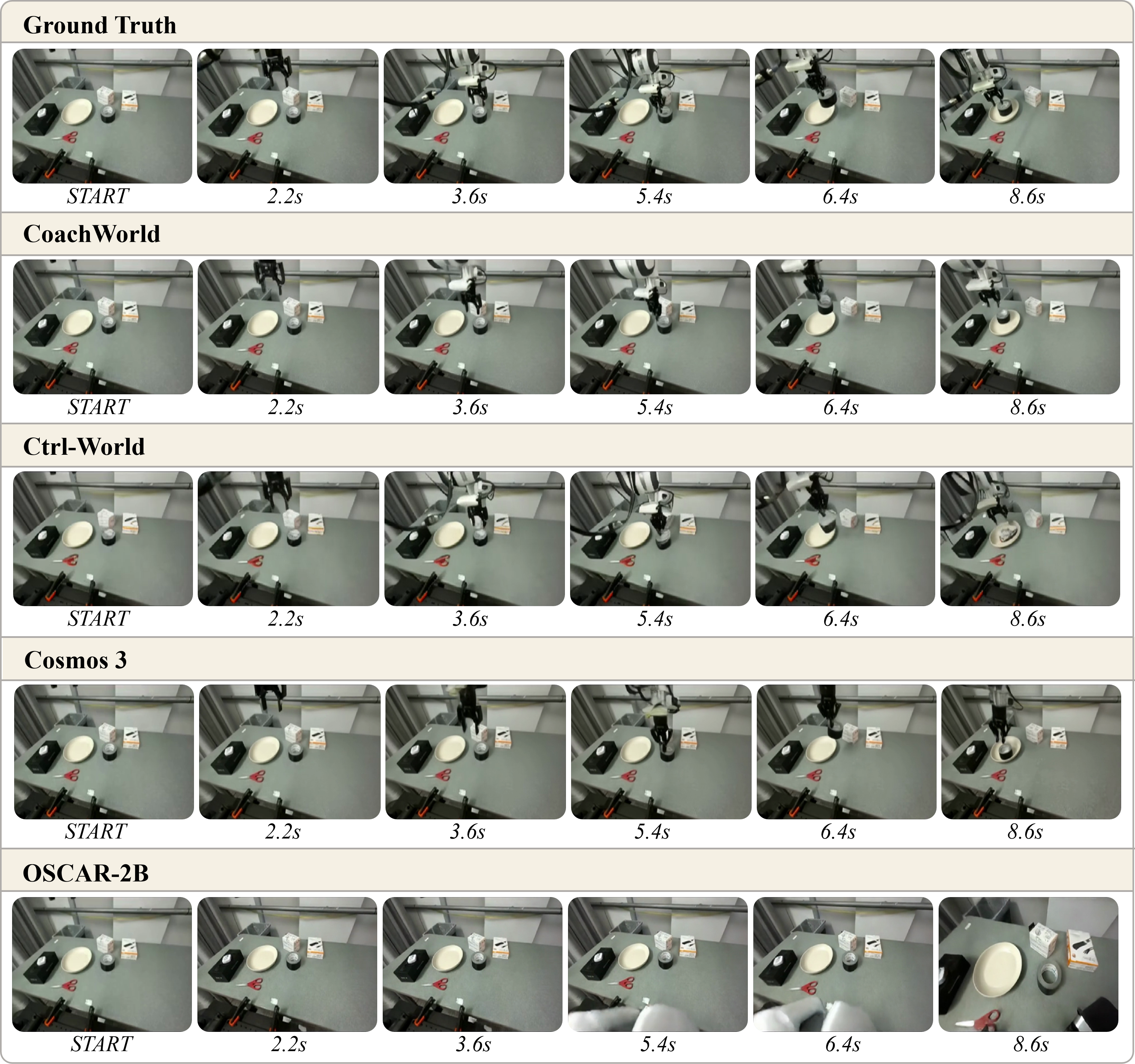}\\[-0.3em]
\textbf{(a)}\par\vspace{0.45em}
\includegraphics[height=0.4\textheight,keepaspectratio]
{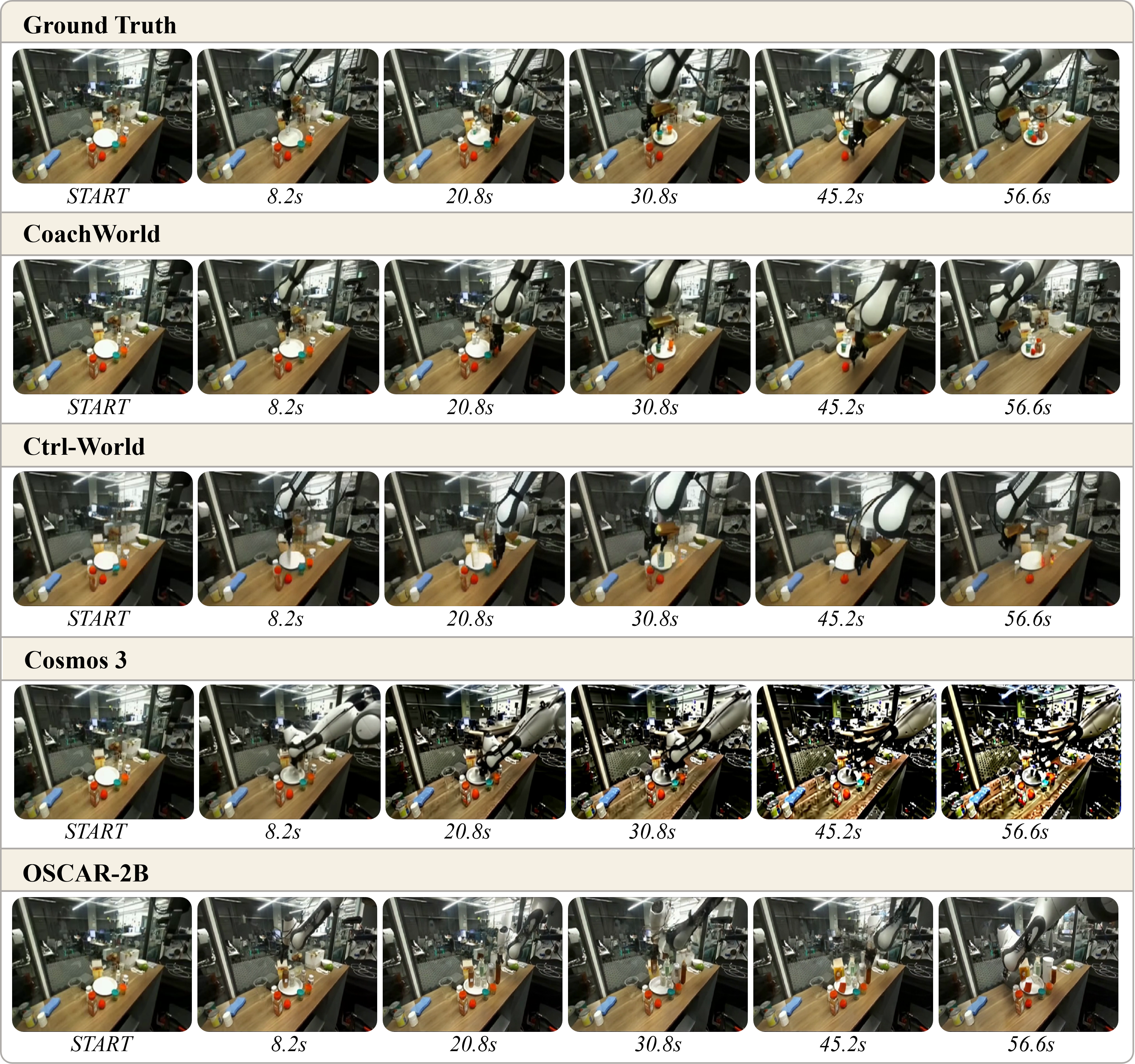}\\[-0.3em]
\textbf{(b)}
\caption{\textbf{Qualitative full-episode prediction.}
Representative episodes shown at matched points on the recorded action clock.}
\label{fig:wm_qualitative}
\end{figure*}

\begin{figure*}[t]
\centering
\includegraphics[width=\textwidth]{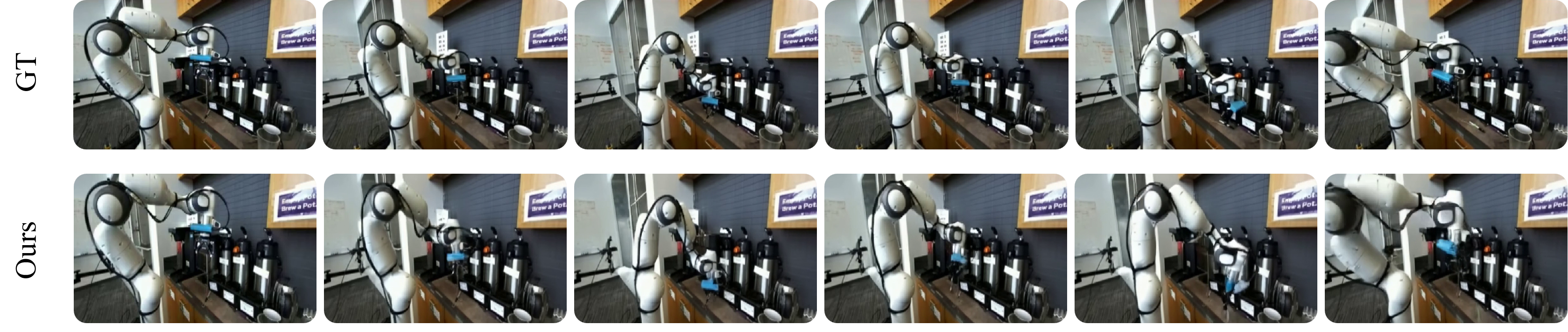}\par\vspace{0.35em}
\textbf{(a) DROID}\par\vspace{0.1em}
\includegraphics[width=\textwidth]{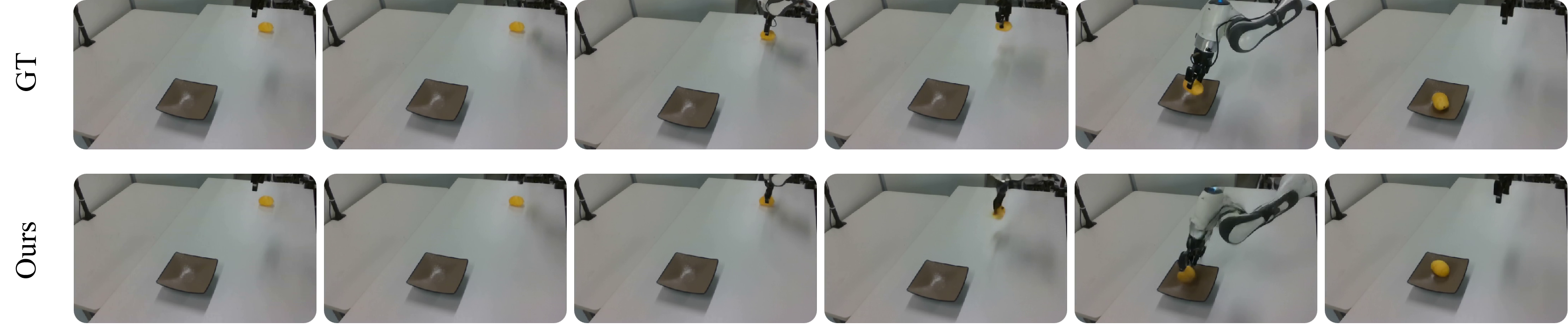}\par\vspace{0.35em}
\textbf{(b) RoboMIND}\par\vspace{0.1em}
\includegraphics[width=\textwidth]{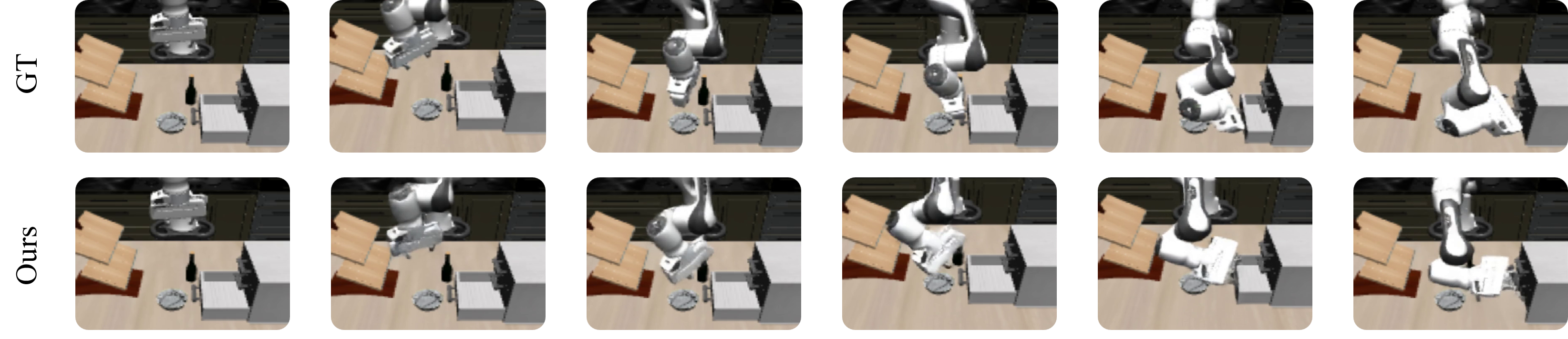}
\textbf{(c) LIBERO}\par\vspace{0.1em}
\caption{\textbf{Representative \modelname{} rollouts on single-arm domains.}
Top rows show reference episodes and bottom rows show \modelname{} predictions; six frames uniformly span each full episode.}
\label{fig:single_arm_dataset_cases}
\end{figure*}

\begin{figure*}[t]
\centering
\includegraphics[width=\textwidth]{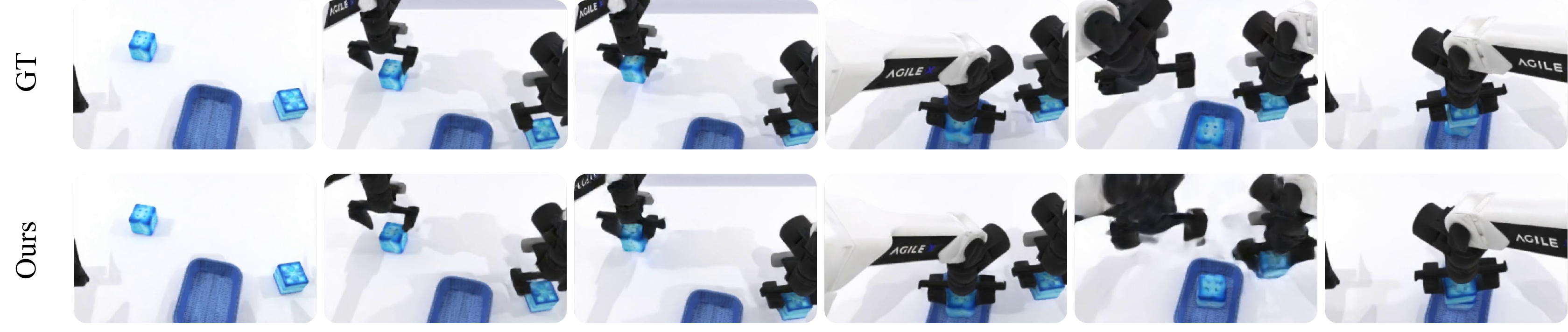}\par\vspace{0.35em}
\textbf{(a) RoboTwin~2.0}\par\vspace{0.1em}
\includegraphics[width=\textwidth]{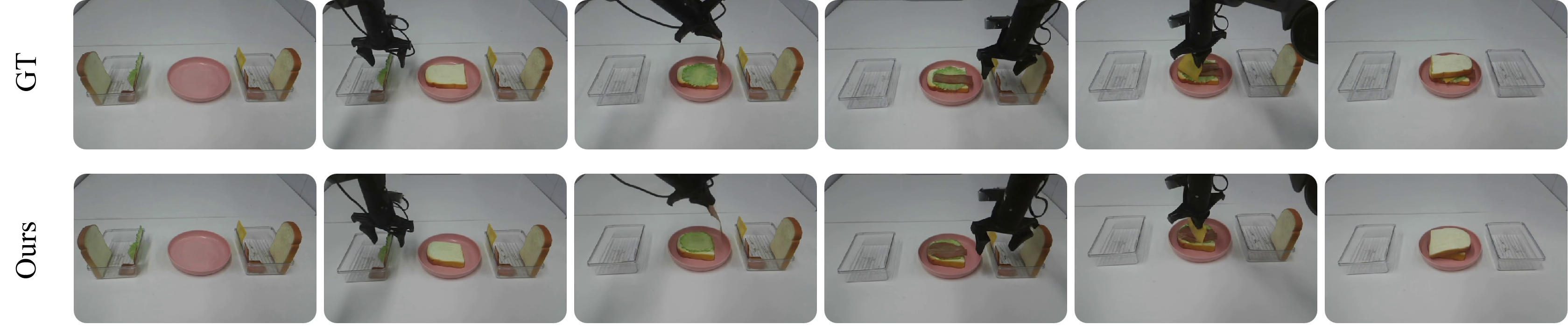}\par\vspace{0.35em}
\textbf{(b) RoboCOIN}\par\vspace{0.1em}
\includegraphics[width=\textwidth]{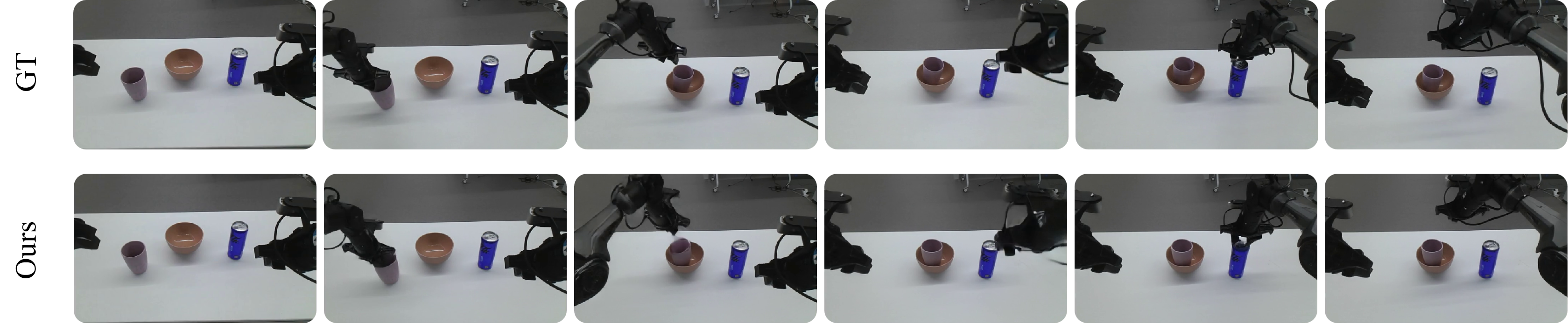}
\textbf{(c) ViFailBack}\par\vspace{0.1em}
\caption{\textbf{Representative \modelname{} rollouts on bimanual domains.}
Top rows show reference episodes and bottom rows show \modelname{} predictions; six frames uniformly span each full episode.}
\label{fig:bimanual_dataset_cases}
\end{figure*}

\FloatBarrier

\subsection{Action Following and EEF-Trajectory Fidelity}
Following EWMBench~\citep{yue2025ewmbenchevaluatingscenemotion}, reference and generated videos are resized to $640\times480$ and processed by the same frozen EEF detector and tracker.
Because both suites contain a single Franka arm, the benchmark's active-hand selection reduces to the unique visible EEF.
The tracker is run independently on prediction and reference; the projected conditioning action is used only for a geometry sanity check and is never treated as the reference-video trajectory.
We report HSD for worst-case spatial deviation, nDTW for temporally ordered path agreement, and DYN for velocity consistency.

\subsection{Blind Human Audit of Physics Adherence and Instruction Following}
We complement the paired visual and trajectory metrics with a blind human audit on fixed, paired subsets of 30 episodes from DROID-180 and 30 episodes from Lab Franka-180.
The same episode subset is used for Ctrl-World, Cosmos~3, OSCAR-2B,
\modelname{} (mixed-domain), \modelname{} (single-domain), and the reference videos.
For each item, annotators receive the task instruction and the complete predicted or reference video.
Method identity and reference status are hidden, and presentation order is randomized.

Two annotators independently assign integer scores from 1 to 5 for Physics Adherence and Instruction Following, without discussion or adjudication.
The rubrics are adapted from MiraBench and WorldArena~\citep{yang2026mirabenchevaluatingactionconditionedreliability,shang2026worldarena}.
Physics Adherence assesses whether visible robot motion, contact, and object response remain physically coherent; Instruction Following assesses whether the video executes the specified task and reaches the requested outcome.
We average scores across annotators and episodes and divide by five to normalize them to $[0,1]$.
The hidden reference videos attain $1.000$ on both dimensions in both domains.
Table~\ref{tab:wm_compact} reports the resulting model scores.

Before adopting the human audit, we piloted an automatic VLM evaluator on the same dimensions.
It failed a reference-ceiling sanity check: some generated rollouts received higher scores than the corresponding ground-truth references.
We therefore do not use the automatic scores in our comparisons.

\FloatBarrier

\subsection{Policy-Success Calibration}
\label{app:policy_success_calibration}
The 22 cells comprise one frozen task--policy-checkpoint pair for each of ten LIBERO-Long, five
RoboTwin, three Franka, and four AgileX tasks. Imagination and deployment use matched prompts,
initial-state strata, and trial counts: 50 trials per task in simulation and 20 on hardware. Each imag-
ined trial averages three generation seeds. This estimates task-level success. The coaching scorecard instead retains modal terminal diagnoses with agreement from at least two seed as mentioned in Section~\ref{sec:diagnosis}.

\FloatBarrier

\section{Router and Diagnostic Pipeline}
\label{app:diagnosis}

\subsection{Router Interface}
At the initial boundary and after each judge-confirmed completion, the router
receives the task instruction, current image, candidate atomic skills, and the
exact completed-subtask prefix.
The deployed prompt is:

\begin{tcblisting}{
  enhanced,
  breakable,
  colback=gray!3,
  colframe=black!50,
  colbacktitle=black!50,
  coltitle=white,
  boxrule=0.45pt,
  arc=2pt,
  left=6pt,
  right=6pt,
  top=5pt,
  bottom=5pt,
  before skip=6pt,
  after skip=6pt,
  title={Language-router prompt},
  fonttitle=\bfseries,
  listing only,
  listing options={
    basicstyle=\ttfamily\scriptsize,
    breaklines=true,
    breakatwhitespace=true,
    breakindent=0pt,
    columns=fullflexible,
    keepspaces=true,
    showstringspaces=false,
    aboveskip=0pt,
    belowskip=0pt
  }
}
SYSTEM
You are a router for a multi-expert robot policy. Given the task,
current observation, candidate atomic skills, and Judge-confirmed
completed subtasks, select the next unfinished atomic subtask.
Treat the completed subtasks as the exact task prefix and never infer
additional completion from the image. Respect the order implied by the
task instruction and copy the selected skill exactly.

Return only {"skill":"...","object":"..."}.

USER
Task: put the bread into the pan, then close the lid
Completed subtasks: pick the bread; place the bread into the pan
Candidate skills: press, pull, pick, place
\end{tcblisting}

The user message also contains the current image.
The expected response is \texttt{\{"skill":"pick",\allowbreak"object":"lid"\}}.
Completion status is supplied exclusively by the judge-confirmed prefix, isolating routing from visual progress estimation.

\subsection{Annotation Bank and Frozen Thresholds}
For each task, the judge test bank contains five successful and five failed reference trajectory clips, all disjoint from judge training and threshold calibration.
Each trajectory stores the canonical route, human expert-switch boundaries, terminal outcome, and failed subtask when applicable.
Two annotators independently label boundaries and failed subtasks, with disagreements resolved by joint review.
Every judge calibrates its per-subtask completion thresholds on the same held-out
calibration bank, after which thresholds and timeouts are frozen.

\subsection{Judge Implementations}
\label{app:judge_baselines}

All judges receive the same causal observation window, task instruction, and
active-subtask instruction, and produce a scalar completion score queried at
the same committed boundaries.
Qwen3-VL uses frozen Qwen3-VL-4B as a zero-shot completion judge.
LIV~\citep{ma2023liv} uses its pretrained language-conditioned visual
representation with domain-specific adaptation for completion prediction.
For Contrastive $\lambda$~\citep{goko2025contrastive}, the original Repformer
representation is unavailable, so we use a frozen CLIP RN50
encoder~\citep{radford2021clip} with a learned temporal completion scorer.
Following SuccessVQA~\citep{du2023successvqa}, we LoRA-fine-tune Qwen3-VL-4B
with binary supervision over completed and incomplete subtask windows.
For TOPReward~\citep{chen2026topreward}, we instantiate its token-scoring
formulation on Qwen3-VL-4B and use the resulting score for completion decisions.
RoboMeter fine-tunes Qwen3-VL-4B to estimate causal subtask progress and
completion.

\subsection{Progress-Judge Evaluation and Metrics}
For Table~\ref{tab:diagnosis}, all scorers receive the same causal reference frames, task text, active-subtask text, and annotated route.
At each committed chunk boundary, the score is compared with the frozen threshold; a crossing switches experts, while no crossing before the frozen deadline
assigns a terminal timeout to the active subtask.
The judge therefore estimates progress rather than explicitly searching for a stall.

On successful trajectories, Switch MAE measures temporal error on matched human transition boundaries, while Switch Recall@$\delta$ measures the fraction of ground-truth transitions recovered within $\delta=1.0$\,s.

Spearman $\rho$ measures rank agreement between predicted and ground-truth
terminal stages over task--stage cells.
Within each evaluation domain, a timeout at stage $k$ is encoded as $k$, while
successful completion of an $n$-stage task is encoded as $n+1$.
We compute $\rho$ separately for LIBERO, RoboTwin, real single-arm, and real
dual-arm settings, and report their unweighted macro average.

For binary transition evaluation, we construct adjacent-stage examples that test
whether the current subtask should be considered complete and control should
advance to the next stage.
A ground-truth transition is treated as the positive class.
TP/TN/FP/FN therefore denote correct and incorrect transition decisions, from
which we compute precision, recall, and F1.
Table~\ref{tab:diagnosis} reports the corresponding results. For Table~\ref{tab:robometer_judge}, switch-time annotations are made
separately on the reference and generated trajectories, and each row
is evaluated against its own annotations. The generated-video result
therefore tests whether the judge detects transitions shown in the
imagined video, not whether those transitions match the real trajectory.

\subsection{Latency and Seed Agreement}
Judge latency is measured at batch size one from submitting the causal frame packet and prompt to receiving a progress score.
On a single NVIDIA A100 80\,GB GPU, p50/p95 latency in seconds is $0.0004/0.0035$ for LIV, $0.0003/0.0034$ for Contrastive $\lambda$, $0.0547/0.3567$ for Qwen3-VL, $0.0488/0.0506$ for SuccessVQA, $0.0663/0.4758$ for TOPReward, and $0.1049/0.1164$ for RoboMeter; world-model generation, decoding, routing, and timeout bookkeeping are excluded.
Each imagined trial uses $R=3$ episode-level world-model seeds.
Their modal task outcome and failed-subtask label determine the scorecard entry; a trial is accepted only when at least two of the three seeds agree.

\begin{table*}[!htbp]
\caption{\textbf{Progress-judge comparison on causal reference-video replay.}
Switch metrics evaluate stage-transition timing, Spearman $\rho$ measures
terminal-stage rank agreement, and outcome metrics evaluate binary transition decisions on adjacent-stage examples.
Best and second-best values are bold and underlined, respectively.}
\label{tab:diagnosis}

\centering
\scriptsize
\setlength{\tabcolsep}{3.0pt}
\renewcommand{\arraystretch}{1.00}

\resizebox{\textwidth}{!}{%
\begin{tabular}{lccccccc}
\toprule

Judge
& Switch MAE (s)$\downarrow$
& Switch Rec.\@1.0s (\%)$\uparrow$
& Spearman $\rho$$\uparrow$
& TP/TN/FP/FN
& Out.-Prec. (\%)$\uparrow$
& Out.-Rec. (\%)$\uparrow$
& Out.-F1 (\%)$\uparrow$ \\
\midrule

LIV$^\dagger$
& 3.311
& 23.64
& 0.328
& 64/67/43/46
& 59.81
& 58.18
& 58.99 \\

Contrastive $\lambda$$^\dagger$
& 1.905
& 28.18
& \underline{0.603}
& 95/80/30/15
& \underline{76.00}
& \underline{86.36}
& \underline{80.85} \\

\midrule

\multicolumn{8}{l}{\emph{VLM-based}} \\

Qwen3-VL
& 0.893
& 20.00
& 0.151
& 29/82/28/81
& 50.88
& 26.36
& 34.73 \\

TOPReward$^\dagger$
& 1.603
& 33.64
& 0.330
& 76/55/55/34
& 58.02
& 69.09
& 63.07 \\

SuccessVQA$^\dagger$
& \underline{0.859}
& \underline{59.09}
& 0.286
& 82/72/38/28
& 68.33
& 74.55
& 71.30 \\

\textbf{RoboMeter}
& \textbf{0.414}
& \textbf{82.73}
& \textbf{0.727}
& 101/92/18/9
& \textbf{84.87}
& \textbf{91.82}
& \textbf{88.21} \\

\bottomrule
\end{tabular}%
}

\vspace{2pt}
\raggedright
\footnotesize
$^\dagger$ denotes an implementation adapted to the shared causal evaluation
setting.

\end{table*}

\FloatBarrier

\section{Coaching Protocol and Per-Round Evidence}
\label{app:coaching}

Algorithm~\ref{alg:robocoach} in the main text summarizes one coaching round.
Here we detail the operator interface, matched acquisition and update
protocols, and per-round results.

\subsection{Operator-Facing Scorecard}
The scorecard displays each subtask--expert pair's imagined success rate, reach count, task-balanced first-timeout mass, and representative failed videos.
For the top $M=2$ pairs, it requests a concrete skill, object or affordance, embodiment, and demonstration count.
The operator checks safety and collects the requested examples without manually reranking the selected pairs.

\begin{figure}[t]
  \centering
  \includegraphics[width=\linewidth]{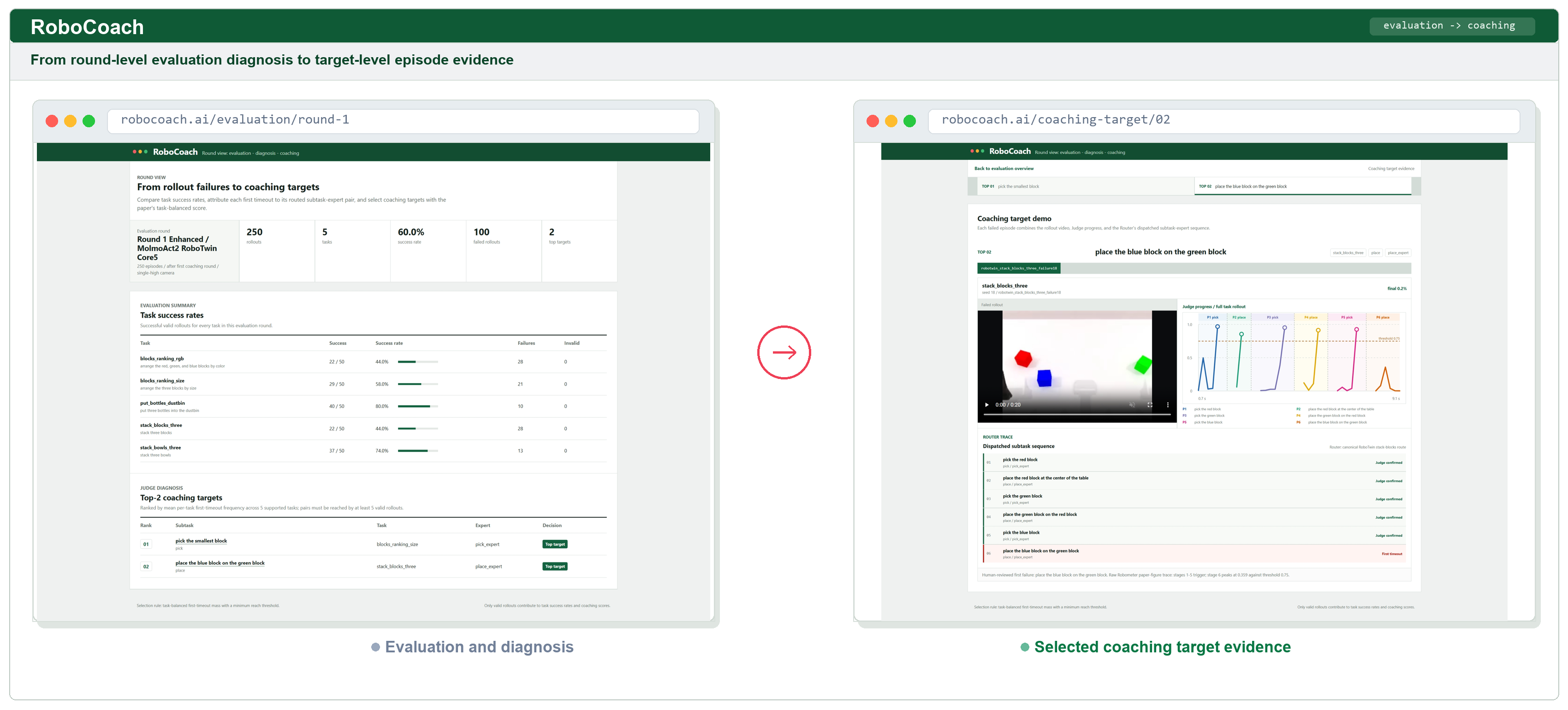}
  \caption{\textbf{From round-level diagnosis to target-level coaching evidence.} Left: \methodname{} aggregates rollout outcomes into task-level success rates and ranks recurring first-timeout subtasks as coaching targets. Right: selecting a target reveals its supporting episode-level evidence, including the failed rollout, judge progress trace, and router-dispatched subtask sequence. This drill-down connects aggregate failure diagnosis to concrete and auditable coaching examples.}
  \label{fig:coaching-evidence-interface}
\end{figure}

Figure~\ref{fig:coaching-evidence-interface} illustrates how \methodname{} turns round-level evaluation results into auditable coaching evidence. The evaluation view aggregates rollout outcomes, diagnoses recurring first-timeout subtasks, and ranks candidate coaching targets. Selecting a target then exposes the corresponding episode-level evidence, including the failed rollout, judge progress trace, and router-dispatched subtask sequence.

\subsection{Matched Coaching Protocol}
Simulation uses three rounds of $B_{\mathrm{sim}}=100$ accepted subtask-level demonstrations and evaluates four methods at cumulative budgets $\{0,100,200,300\}$:
Single VLA + Uniform, Single VLA + WM-targeted, Modular + Random, and \methodname{}.
The two Single-VLA methods update one shared global LoRA adapter using all acquired data.
Modular + Random randomly selects two subtask--expert pairs per round,
whereas \methodname{} selects the top two pairs from the world-model
scorecard. Both modular conditions update the experts associated with the
selected pairs; when two selected pairs share an expert, their
demonstrations are pooled for the same update.
Single VLA + WM-targeted and \methodname{} receive the same scorecard-requested demonstrations, isolating shared-global versus selected-expert updates on targeted data.
All coaching conditions start from the same initial policy checkpoint.
For the modular conditions, each skill expert is initialized from the same
initial adapter before coaching begins.
All conditions use the same router and progress judge; the Single-VLA
conditions apply the routed subtask instruction to one shared global adapter.
Accordingly, SR@0 is a shared evaluation of the common initialization, and the
conditions diverge only after coaching begins.
Hardware uses $B_{\mathrm{real}}=50$ and compares Single VLA + Uniform with \methodname{} at $\{0,50,100,150\}$.
Within each domain, methods use the same demonstration budget and evaluation-trial protocol.

For Single VLA, one shared LoRA adapter is updated on all acquired data.
For the modular conditions, only the selected expert adapters are updated.

To quantify uncertainty in Figure~\ref{fig:coaching_curve}, we hold the
benchmark tasks fixed and resample evaluation outcomes within each task.
For each method and checkpoint, we generate 50,000 bootstrap replicates
by drawing each task's success count from a binomial distribution with
its observed success fraction and original trial count (50 for simulation
or 20 for real robots). In each replicate, we average task success rates
equally. The 2.5th and 97.5th percentiles define a pointwise 95\%
confidence interval.

For task-macro success $S_q$ after round $q$, normalized Budget AUC is
\begin{equation}
    \mathrm{AUC}_{0:3B}=\frac{S_0+2S_1+2S_2+S_3}{6}.
\end{equation}
\FloatBarrier
\subsection{Acquisition Trace and Per-Round Results}

\begin{table*}[h]
\caption{\textbf{Per-round acquisition trace.}
\methodname{} rows report the two scorecard-selected subtask--expert targets in each round,
while Random rows report the two randomly selected targets.
Single VLA + WM-targeted shares the \methodname{} acquisitions; Uniform has no targeted selection.}
\label{tab:allocation_trace}
\centering
\scriptsize
\setlength{\tabcolsep}{3.5pt}
\renewcommand{\arraystretch}{1.10}
\resizebox{\textwidth}{!}{%
\begin{tabular}{llp{3.8cm}p{3.8cm}p{3.8cm}}
\toprule
Domain & Method & Round 1 & Round 2 & Round 3 \\
\midrule

LIBERO
& Random
& \acquisitioncell{
L6: pick white mug $\rightarrow$ pick\\
L8: place first moka pot $\rightarrow$ place}
& \acquisitioncell{
L0: pick tomato sauce $\rightarrow$ pick\\
L7: place alphabet soup $\rightarrow$ place}
& \acquisitioncell{
L7: pick cream-cheese box $\rightarrow$ pick\\
L9: place mug in\\
microwave $\rightarrow$ place}
\\

\cmidrule(lr){1-5}

LIBERO
& \methodname{}
& \acquisitioncell{
L4: pick white mug $\rightarrow$ pick\\
L6: pick chocolate pudding $\rightarrow$ pick}
& \acquisitioncell{
L0: pick soup / tomato sauce $\rightarrow$ pick\\
L8: pick first / second moka pot $\rightarrow$ pick}
& \acquisitioncell{
L8: place first moka pot $\rightarrow$ place\\
L9: place mug in microwave $\rightarrow$ place}
\\

\midrule

RoboTwin
& Random
& \acquisitioncell{
R1: pick largest block $\rightarrow$ pick\\
R3: pick third bowl $\rightarrow$ pick}
& \acquisitioncell{
R0: pick blue block $\rightarrow$ pick\\
R3: place first bowl $\rightarrow$ place}
& \acquisitioncell{
R3: pick second bowl $\rightarrow$ pick\\
R4: place second block on first $\rightarrow$ place}
\\

\cmidrule(lr){1-5}

RoboTwin
& \methodname{}
& \acquisitioncell{
R1: place medium block $\rightarrow$ place\\
R2: pick first bottle $\rightarrow$ pick}
& \acquisitioncell{
R0: pick green block $\rightarrow$ pick\\
R1: place largest block $\rightarrow$ place}
& \acquisitioncell{
R4: place third block on\\
second $\rightarrow$ place\\
R3: place third bowl on\\
second $\rightarrow$ place}
\\

\midrule

Franka
& \methodname{}
& \acquisitioncell{
pull lamp cord $\rightarrow$ pull\\
place lid $\rightarrow$ place}
& \acquisitioncell{
press green button $\rightarrow$ press\\
place lid $\rightarrow$ place}
& \acquisitioncell{
pull lamp cord $\rightarrow$ pull\\
press red button $\rightarrow$ press}
\\

\midrule

AgileX
& \methodname{}
& \acquisitioncell{
pick right shoe $\rightarrow$ pick\\
place tea bag $\rightarrow$ place}
& \acquisitioncell{
pick red block $\rightarrow$ pick\\
push plate $\rightarrow$ push}
& \acquisitioncell{
pick red block $\rightarrow$ pick\\
place left shoe $\rightarrow$ place}
\\

\bottomrule
\end{tabular}%
}
\end{table*}

\begin{table*}[!htbp]
\caption{\textbf{Per-round complete-task success.}
$B=100$ in simulation and $B=50$ on hardware.}
\label{tab:coaching_per_round}
\centering
\scriptsize
\setlength{\tabcolsep}{5pt}
\begin{tabular}{llccccc}
\toprule
Domain & Method & SR@0 & SR@B & SR@2B & SR@3B & Budget AUC \\
\midrule
LIBERO & Single VLA + Uniform & 66.0\%&66.8\%&66.4\%&67.0\%&66.6\% \\
LIBERO & Single VLA + WM-targeted & 66.0\%&68.2\%&66.8\%&67.8\%&67.3\% \\
LIBERO & Modular + Random & 66.0\%&67.8\%&69.2\%&68.4\%&68.1\% \\
LIBERO & \methodname{} & 66.0\%&69.6\%&70.6\%&71.2\%&69.6\% \\

RoboTwin & Single VLA + Uniform
& 58.8\% & 58.4\% & 58.0\% & 59.6\% & 58.5\% \\
RoboTwin & Single VLA + WM-targeted
& 58.8\% & 54.8\% & 57.2\% & 54.8\% & 56.3\% \\
RoboTwin & Modular + Random
& 58.8\% & 58.8\% & 56.4\% & 55.2\% & 57.4\% \\
RoboTwin & \methodname{}
& 58.8\% & 64.0\% & 65.2\% & 68.0\% & 64.2\% \\

\midrule
Franka & Single VLA + Uniform & 13.3\%&15.0\%&20.0\%&30.0\%&18.9\% \\
Franka & \methodname{} & 13.3\%&46.7\%&68.3\%&75.0\%&53.1\% \\
AgileX & Single VLA + Uniform & 40.0\%&48.8\%&46.3\%&47.5\%&46.3\% \\
AgileX & \methodname{} & 40.0\%&72.5\%&83.8\%&83.8\%&72.7\% \\
\bottomrule
\end{tabular}
\end{table*}

\FloatBarrier
\subsection{LIBERO Per-Task Coaching Results}
\label{app:libero_per_task}

Table~\ref{tab:libero_per_task_coaching} reports the complete per-task
LIBERO results underlying the task-macro curves in
Table~\ref{tab:coaching_per_round}.
All methods use the same frozen SR@0 evaluation, yielding 330/500 successes
(66.0\%). Each task--round cell contains 50 evaluation episodes under the same
fixed-seed protocol. Cells corresponding to updated global LoRAs or routed
experts use their respective frozen checkpoints; non-updated modular cells are
independently re-evaluated rather than reusing outcomes from another method or
round.

\begin{table*}[!htbp]
\caption{\textbf{Per-task LIBERO coaching results under the fixed evaluation protocol.}
Task IDs follow the LIBERO-Long mapping in Table~\ref{tab:libero_long_tasks}.
Each cell reports successes out of 50 fixed-seed evaluation episodes, with the corresponding success rate in parentheses.
All methods share the same SR@0 evaluation of 330/500 (66.0\%).}
\label{tab:libero_per_task_coaching}
\centering
\scriptsize
\setlength{\tabcolsep}{6pt}
\renewcommand{\arraystretch}{0.96}

\begin{tabular}{lcccc}
\toprule
Task ID & SR@0 & SR@B & SR@2B & SR@3B \\
\midrule

\multicolumn{5}{l}{\emph{Single VLA + Uniform}} \\
L0 & 25/50 (50\%) & 29/50 (58\%) & 30/50 (60\%) & 25/50 (50\%) \\
L1 & 30/50 (60\%) & 35/50 (70\%) & 32/50 (64\%) & 36/50 (72\%) \\
L2 & 39/50 (78\%) & 38/50 (76\%) & 35/50 (70\%) & 35/50 (70\%) \\
L3 & 43/50 (86\%) & 45/50 (90\%) & 45/50 (90\%) & 44/50 (88\%) \\
L4 & 33/50 (66\%) & 26/50 (52\%) & 29/50 (58\%) & 30/50 (60\%) \\
L5 & 41/50 (82\%) & 42/50 (84\%) & 43/50 (86\%) & 44/50 (88\%) \\
L6 & 29/50 (58\%) & 32/50 (64\%) & 30/50 (60\%) & 35/50 (70\%) \\
L7 & 38/50 (76\%) & 35/50 (70\%) & 36/50 (72\%) & 28/50 (56\%) \\
L8 & 17/50 (34\%) & 22/50 (44\%) & 20/50 (40\%) & 28/50 (56\%) \\
L9 & 35/50 (70\%) & 30/50 (60\%) & 32/50 (64\%) & 30/50 (60\%) \\
\cmidrule(lr){1-5}
\textbf{Total}
& \textbf{330/500 (66.0\%)}
& \textbf{334/500 (66.8\%)}
& \textbf{332/500 (66.4\%)}
& \textbf{335/500 (67.0\%)} \\

\midrule
\multicolumn{5}{l}{\emph{Single VLA + WM-targeted}} \\
L0 & 25/50 (50\%) & 26/50 (52\%) & 29/50 (58\%) & 30/50 (60\%) \\
L1 & 30/50 (60\%) & 35/50 (70\%) & 35/50 (70\%) & 36/50 (72\%) \\
L2 & 39/50 (78\%) & 35/50 (70\%) & 39/50 (78\%) & 39/50 (78\%) \\
L3 & 43/50 (86\%) & 46/50 (92\%) & 46/50 (92\%) & 45/50 (90\%) \\
L4 & 33/50 (66\%) & 28/50 (56\%) & 29/50 (58\%) & 34/50 (68\%) \\
L5 & 41/50 (82\%) & 44/50 (88\%) & 44/50 (88\%) & 43/50 (86\%) \\
L6 & 29/50 (58\%) & 30/50 (60\%) & 27/50 (54\%) & 31/50 (62\%) \\
L7 & 38/50 (76\%) & 35/50 (70\%) & 34/50 (68\%) & 34/50 (68\%) \\
L8 & 17/50 (34\%) & 24/50 (48\%) & 23/50 (46\%) & 19/50 (38\%) \\
L9 & 35/50 (70\%) & 38/50 (76\%) & 28/50 (56\%) & 28/50 (56\%) \\
\cmidrule(lr){1-5}
\textbf{Total}
& \textbf{330/500 (66.0\%)}
& \textbf{341/500 (68.2\%)}
& \textbf{334/500 (66.8\%)}
& \textbf{339/500 (67.8\%)} \\

\midrule
\multicolumn{5}{l}{\emph{Modular + Random}} \\
L0 & 25/50 (50\%) & 25/50 (50\%) & 30/50 (60\%) & 33/50 (66\%) \\
L1 & 30/50 (60\%) & 30/50 (60\%) & 30/50 (60\%) & 30/50 (60\%) \\
L2 & 39/50 (78\%) & 38/50 (76\%) & 40/50 (80\%) & 39/50 (78\%) \\
L3 & 43/50 (86\%) & 43/50 (86\%) & 43/50 (86\%) & 43/50 (86\%) \\
L4 & 33/50 (66\%) & 32/50 (64\%) & 32/50 (64\%) & 32/50 (64\%) \\
L5 & 41/50 (82\%) & 41/50 (82\%) & 41/50 (82\%) & 41/50 (82\%) \\
L6 & 29/50 (58\%) & 34/50 (68\%) & 34/50 (68\%) & 32/50 (64\%) \\
L7 & 38/50 (76\%) & 38/50 (76\%) & 37/50 (74\%) & 36/50 (72\%) \\
L8 & 17/50 (34\%) & 23/50 (46\%) & 23/50 (46\%) & 29/50 (58\%) \\
L9 & 35/50 (70\%) & 35/50 (70\%) & 36/50 (72\%) & 27/50 (54\%) \\
\cmidrule(lr){1-5}
\textbf{Total}
& \textbf{330/500 (66.0\%)}
& \textbf{339/500 (67.8\%)}
& \textbf{346/500 (69.2\%)}
& \textbf{342/500 (68.4\%)} \\

\midrule
\multicolumn{5}{l}{\textbf{\methodname{} (ours)}} \\
L0 & 25/50 (50\%) & 25/50 (50\%) & 31/50 (62\%) & 31/50 (62\%) \\
L1 & 30/50 (60\%) & 35/50 (70\%) & 30/50 (60\%) & 30/50 (60\%) \\
L2 & 39/50 (78\%) & 39/50 (78\%) & 38/50 (76\%) & 40/50 (80\%) \\
L3 & 43/50 (86\%) & 43/50 (86\%) & 43/50 (86\%) & 43/50 (86\%) \\
L4 & 33/50 (66\%) & 33/50 (66\%) & 35/50 (70\%) & 35/50 (70\%) \\
L5 & 41/50 (82\%) & 46/50 (92\%) & 41/50 (82\%) & 41/50 (82\%) \\
L6 & 29/50 (58\%) & 36/50 (72\%) & 35/50 (70\%) & 35/50 (70\%) \\
L7 & 38/50 (76\%) & 38/50 (76\%) & 38/50 (76\%) & 38/50 (76\%) \\
L8 & 17/50 (34\%) & 18/50 (36\%) & 27/50 (54\%) & 30/50 (60\%) \\
L9 & 35/50 (70\%) & 35/50 (70\%) & 35/50 (70\%) & 33/50 (66\%) \\
\cmidrule(lr){1-5}
\textbf{Total}
& \textbf{330/500 (66.0\%)}
& \textbf{348/500 (69.6\%)}
& \textbf{353/500 (70.6\%)}
& \textbf{356/500 (71.2\%)} \\

\bottomrule
\end{tabular}
\end{table*}

\FloatBarrier
\subsection{RoboTwin Per-Task Coaching Results}
\label{app:robotwin_per_task}

Table~\ref{tab:robotwin_per_task_coaching} reports the complete per-task
RoboTwin~2.0 results underlying the task-macro coaching curves.
Task IDs R0--R4 follow the mapping in Table~\ref{tab:robotwin_tasks}.
Each task is evaluated over 50 trials at every coaching checkpoint.

\begin{table*}[!htbp]
\caption{\textbf{Per-task RoboTwin~2.0 coaching results under the fixed evaluation protocol.}
Task IDs follow the RoboTwin task mapping in Table~\ref{tab:robotwin_tasks}.
Each cell reports successes out of 50 fixed-seed evaluation episodes, with the
corresponding success rate in parentheses.
All methods share the same SR@0 evaluation of 147/250 (58.8\%).}
\label{tab:robotwin_per_task_coaching}
\centering
\scriptsize
\setlength{\tabcolsep}{5.0pt}
\renewcommand{\arraystretch}{1.08}

\begin{tabular}{lcccc}
\toprule
Task ID & SR@0 & SR@B & SR@2B & SR@3B \\
\midrule

\multicolumn{5}{l}{\emph{Single VLA + Uniform}} \\
R0
& 25/50 (50\%)
& 26/50 (52\%)
& 25/50 (50\%)
& 25/50 (50\%) \\
R1
& 22/50 (44\%)
& 22/50 (44\%)
& 20/50 (40\%)
& 24/50 (48\%) \\
R2
& 33/50 (66\%)
& 41/50 (82\%)
& 38/50 (76\%)
& 37/50 (74\%) \\
R3
& 39/50 (78\%)
& 37/50 (74\%)
& 43/50 (86\%)
& 41/50 (82\%) \\
R4
& 28/50 (56\%)
& 20/50 (40\%)
& 19/50 (38\%)
& 22/50 (44\%) \\
\cmidrule(lr){1-5}
\textbf{Total}
& \textbf{147/250 (58.8\%)}
& \textbf{146/250 (58.4\%)}
& \textbf{145/250 (58.0\%)}
& \textbf{149/250 (59.6\%)} \\

\midrule
\multicolumn{5}{l}{\emph{Single VLA + WM-targeted}} \\
R0
& 25/50 (50\%)
& 25/50 (50\%)
& 23/50 (46\%)
& 24/50 (48\%) \\
R1
& 22/50 (44\%)
& 20/50 (40\%)
& 26/50 (52\%)
& 23/50 (46\%) \\
R2
& 33/50 (66\%)
& 34/50 (68\%)
& 34/50 (68\%)
& 33/50 (66\%) \\
R3
& 39/50 (78\%)
& 37/50 (74\%)
& 38/50 (76\%)
& 37/50 (74\%) \\
R4
& 28/50 (56\%)
& 21/50 (42\%)
& 22/50 (44\%)
& 20/50 (40\%) \\
\cmidrule(lr){1-5}
\textbf{Total}
& \textbf{147/250 (58.8\%)}
& \textbf{137/250 (54.8\%)}
& \textbf{143/250 (57.2\%)}
& \textbf{137/250 (54.8\%)} \\

\midrule
\multicolumn{5}{l}{\emph{Modular + Random}} \\
R0
& 25/50 (50\%)
& 27/50 (54\%)
& 25/50 (50\%)
& 26/50 (52\%) \\
R1
& 22/50 (44\%)
& 23/50 (46\%)
& 19/50 (38\%)
& 22/50 (44\%) \\
R2
& 33/50 (66\%)
& 32/50 (64\%)
& 32/50 (64\%)
& 33/50 (66\%) \\
R3
& 39/50 (78\%)
& 38/50 (76\%)
& 38/50 (76\%)
& 34/50 (68\%) \\
R4
& 28/50 (56\%)
& 27/50 (54\%)
& 27/50 (54\%)
& 23/50 (46\%) \\
\cmidrule(lr){1-5}
\textbf{Total}
& \textbf{147/250 (58.8\%)}
& \textbf{147/250 (58.8\%)}
& \textbf{141/250 (56.4\%)}
& \textbf{138/250 (55.2\%)} \\

\midrule
\multicolumn{5}{l}{\textbf{\methodname{} (ours)}} \\
R0
& 25/50 (50\%)
& 25/50 (50\%)
& 26/50 (52\%)
& 26/50 (52\%) \\
R1
& 22/50 (44\%)
& 26/50 (52\%)
& 26/50 (52\%)
& 25/50 (50\%) \\
R2
& 33/50 (66\%)
& 40/50 (80\%)
& 47/50 (94\%)
& 47/50 (94\%) \\
R3
& 39/50 (78\%)
& 41/50 (82\%)
& 36/50 (72\%)
& 44/50 (88\%) \\
R4
& 28/50 (56\%)
& 28/50 (56\%)
& 28/50 (56\%)
& 28/50 (56\%) \\
\cmidrule(lr){1-5}
\textbf{Total}
& \textbf{147/250 (58.8\%)}
& \textbf{160/250 (64.0\%)}
& \textbf{163/250 (65.2\%)}
& \textbf{170/250 (68.0\%)} \\

\bottomrule
\end{tabular}
\end{table*}

\FloatBarrier
\subsection{Franka Per-Task Coaching Results}
\label{app:franka_per_task}

Table~\ref{tab:franka_per_task_coaching} reports the per-task Franka results
across the three coaching rounds. Here, $B=50$ accepted demonstrations per round.

\begin{table}[!htbp]
\caption{\textbf{Per-task coaching results on Franka.}
Each entry reports successes out of 20 evaluation trials, with success rate in parentheses.}
\label{tab:franka_per_task_coaching}
\centering
\scriptsize
\setlength{\tabcolsep}{4.5pt}
\renewcommand{\arraystretch}{1.05}

\begin{tabular}{lcccc}
\toprule
Task & SR@0 & SR@B & SR@2B & SR@3B \\
\midrule

\multicolumn{5}{l}{\emph{Single VLA + Uniform}} \\
Lamp switch \& cord
& 3/20 (15\%) & 6/20 (30\%) & 2/20 (10\%) & 8/20 (40\%) \\
Two buttons
& 3/20 (15\%) & 2/20 (10\%) & 3/20 (15\%) & 4/20 (20\%) \\
Bread \& pan
& 2/20 (10\%) & 1/20 (5\%) & 7/20 (35\%) & 6/20 (30\%) \\
\cmidrule(lr){1-5}
\textbf{Total}
& \textbf{8/60 (13.3\%)}
& \textbf{9/60 (15.0\%)}
& \textbf{12/60 (20.0\%)}
& \textbf{18/60 (30.0\%)} \\

\midrule
\multicolumn{5}{l}{\textbf{\methodname{} (ours)}} \\
Lamp switch \& cord
& 3/20 (15\%) & 14/20 (70\%) & 14/20 (70\%) & 17/20 (85\%) \\
Two buttons
& 3/20 (15\%) & 3/20 (15\%) & 13/20 (65\%) & 14/20 (70\%) \\
Bread \& pan
& 2/20 (10\%) & 11/20 (55\%) & 14/20 (70\%) & 14/20 (70\%) \\
\cmidrule(lr){1-5}
\textbf{Total}
& \textbf{8/60 (13.3\%)}
& \textbf{28/60 (46.7\%)}
& \textbf{41/60 (68.3\%)}
& \textbf{45/60 (75.0\%)} \\

\bottomrule
\end{tabular}
\end{table}

\FloatBarrier
\subsection{AgileX Per-Task Coaching Results}
\label{app:agilex_per_task}

Table~\ref{tab:agilex_per_task_coaching} reports the per-task AgileX results
across the three coaching rounds. Here, $B=50$ accepted demonstrations per round.

\begin{table}[!htbp]
\caption{\textbf{Per-task coaching results on AgileX.}
Each entry reports successes out of 20 evaluation trials, with success rate in parentheses.}
\label{tab:agilex_per_task_coaching}
\centering
\scriptsize
\setlength{\tabcolsep}{4.5pt}
\renewcommand{\arraystretch}{1.05}

\begin{tabular}{lcccc}
\toprule
Task & SR@0 & SR@B & SR@2B & SR@3B \\
\midrule

\multicolumn{5}{l}{\emph{Single VLA + Uniform}} \\
Block \& drawer
& 12/20 (60\%) & 14/20 (70\%) & 14/20 (70\%) & 12/20 (60\%) \\
Shoes \& box
& 5/20 (25\%) & 6/20 (30\%) & 4/20 (20\%) & 5/20 (25\%) \\
Table setting
& 13/20 (65\%) & 16/20 (80\%) & 17/20 (85\%) & 17/20 (85\%) \\
Tea making
& 2/20 (10\%) & 3/20 (15\%) & 2/20 (10\%) & 4/20 (20\%) \\
\cmidrule(lr){1-5}
\textbf{Total}
& \textbf{32/80 (40.0\%)}
& \textbf{39/80 (48.8\%)}
& \textbf{37/80 (46.3\%)}
& \textbf{38/80 (47.5\%)} \\

\midrule
\multicolumn{5}{l}{\textbf{\methodname{} (ours)}} \\
Block \& drawer
& 12/20 (60\%) & 12/20 (60\%) & 15/20 (75\%) & 17/20 (85\%) \\
Shoes \& box
& 5/20 (25\%) & 14/20 (70\%) & 14/20 (70\%) & 12/20 (60\%) \\
Table setting
& 13/20 (65\%) & 13/20 (65\%) & 20/20 (100\%) & 20/20 (100\%) \\
Tea making
& 2/20 (10\%) & 19/20 (95\%) & 18/20 (90\%) & 18/20 (90\%) \\
\cmidrule(lr){1-5}
\textbf{Total}
& \textbf{32/80 (40.0\%)}
& \textbf{58/80 (72.5\%)}
& \textbf{67/80 (83.8\%)}
& \textbf{67/80 (83.8\%)} \\

\bottomrule
\end{tabular}
\end{table}

\FloatBarrier
\section{Held-Out Composition Protocol}
\label{app:composition}

The four held-out cases correspond to the five-stage AgileX continuation
(Case A), the Franka button-to-lamp cross-task composition (Case B),
AgileX shoe reordering (Case C), and Franka button reordering (Case D).
For every case, all constituent atomic skills are supported by existing experts,
while the complete ordered route and its semantic paraphrases are excluded from
policy and judge training, coaching data, router examples, threshold calibration,
and checkpoint selection.
The evaluation therefore isolates the reuse of previously learned skills along
unseen task routes.

\paragraph{Route descriptions.}
Case A combines the table-setting and tea-making skills into the route
\emph{wipe the table} $\rightarrow$ \emph{push the plate to the center}
$\rightarrow$ \emph{place the cup on the plate}
$\rightarrow$ \emph{place the tea bag into the cup}
$\rightarrow$ \emph{pour water into the cup}. Case B combines the green-button
skill from \emph{Two buttons press} with the lamp-cord skill from \emph{Lamp
switch and cord}. Case C reverses the right- and left-shoe placement skills from
\emph{Shoes and box} before closing the box. Case D reverses the two-button
order from \emph{Two buttons press}, pressing red before green and reusing the
same press expert with a different argument.

\begin{table}[!htbp]
\centering
\caption{\textbf{Held-out composition success by case.}
Task-macro success is the arithmetic mean of the four case-level rates.}
\label{tab:heldout_composition_counts}
\small
\setlength{\tabcolsep}{4pt}
\begin{tabular}{lccc}
\toprule
Case & Trials & \methodname{} & Single VLA + Uniform \\
\midrule
A: continuation & 20 & 13/20 (65\%) & 0/20 (0\%) \\
B: composition & 20 & 3/20 (15\%) & 0/20 (0\%) \\
C: AgileX reordering & 20 & 5/20 (25\%) & 0/20 (0\%) \\
D: Franka reordering & 20 & 7/20 (35\%) & 0/20 (0\%) \\
\midrule
Task macro & -- & 35.0\% & 0.0\% \\
\bottomrule
\end{tabular}
\end{table}

\end{document}